\documentclass{article}

\usepackage[utf8]{inputenc} 
\usepackage[T1]{fontenc}    
\usepackage{hyperref}       
\usepackage{url}            
\usepackage{booktabs}       
\usepackage{amsfonts}       
\usepackage{nicefrac}       
\usepackage{color}
\usepackage{graphicx}
\usepackage{natbib}
\usepackage[margin=1.3in]{geometry}
\usepackage{multirow}
\usepackage{times}
\usepackage{mathptmx} 
\usepackage{bm} 
\usepackage{amsmath,amsthm,amssymb}
\usepackage{wrapfig}
 \usepackage{subfigure}
\usepackage[table]{xcolor}
\newtheorem{theorem}{Theorem}
\newtheorem{proposition}{Proposition}

\newtheorem{corollary}{Corollary}

\newcommand{\bl[1]}{{\bf #1}}
\newcommand{\bz}{\bl[z]}
\newcommand{\bxi}{\bm{\xi}}
\newcommand{\bgamma}{\bm{\gamma}}
\newcommand{\bx}{{\bf x}}

\newcommand{\Domega}{D_\Omega}
\newcommand{\bv}{\bl[v]}
\newcommand{\bu}{\bl[u]}

\begin{document}
\title{Preferential Bayesian optimisation with Skew Gaussian Processes }

\author{Alessio~Benavoli,
        Dario~Azzimonti,
        and~Dario~Piga
\thanks{Alessio Benavoli is at
School of Computer Science and Statistics, Trinity College, Ireland
              email: \text{alessio.benavoli@tcd.ie}  }         
\thanks{ Dario Azzimonti and Dario Piga are at
           Dalle Molle Institute for Artificial Intelligence Research (IDSIA) - USI/SUPSI, Manno, Switzerland. }}


\maketitle

\begin{abstract}
Bayesian optimisation (BO) is a very effective approach for sequential black-box
optimization where direct queries of the objective function are expensive. However, there are cases where the objective function can only be accessed via preference judgments, such as "this is better than that" between two candidate solutions (like in A/B tests or recommender systems). The state-of-the-art approach to Preferential Bayesian Optimization (PBO) uses a Gaussian process to model the preference function and a Bernoulli likelihood to model the observed pairwise comparisons. Laplace’s method is then employed to compute posterior inferences and, in particular, to build an appropriate acquisition function.  
In this paper, we prove that the true posterior distribution of the preference function is a Skew Gaussian Process (SkewGP), with highly skewed pairwise marginals and, thus, show that Laplace's method usually provides a very poor approximation. We then derive an efficient method to compute the exact SkewGP posterior and use it as surrogate model for PBO employing standard acquisition functions (Upper Credible Bound, etc.). We illustrate the benefits of our exact PBO-SkewGP in a variety of experiments, by showing that it consistently outperforms PBO based on Laplace's approximation both in terms of convergence speed and computational time. We also show that our framework can be extended to deal with mixed preferential-categorical BO, typical for instance in smart manufacturing, where binary judgments (valid or non-valid) together with preference judgments are available.	
\end{abstract}



\section{Introduction}
Bayesian optimization (BO) is a powerful tool for global optimisation of expensive-to-evaluate black-box objective functions~\citep{BCD10,mockus2012bayesian}. However, in many realistic scenarios, the objective function to be optimized  cannot be easily quantified. This happens for instance in optimizing 
chemical and manufacturing processes, in cases where judging  the quality of the final product  can be a difficult and costly task, or simply in situations where only human preferences are available, like in A/B tests~\citep{siroker2013b}.  
  In such situations,  Preferential Bayesian optimization (PBO) \citep{gonzalez2017preferential} or more general algorithms for active preference learning should be adopted~\citep{BDG08,pmlr-v32-zoghi14,sadigh2017active,bemporad2019active}. These approaches require the users to  simply  compare the final outcomes of two different experiments  and indicate which they prefer. Indeed, it is well known that humans are better at comparing  two options rather than assessing the value of ``goodness'' of an  option~\citep{Thu27,CBG15}. 
 
This contribution is focused on PBO. As in state-of-the-art approach for PBO~\citep{gonzalez2017preferential}, we use a Gaussian process (GP) as a prior distribution of the \emph{latent  preference function} and a probit  likelihood to model the observed pairwise comparisons. However, our contribution differs from and  improves~\cite{gonzalez2017preferential} in several directions. 

First, the state-of-the-art PBO methods usually approximate the posterior distribution of the preference function via Laplace's approach. On the other hand, we  compute  the exact  posterior distribution,  which we prove  to be  a Skew Gaussian Process (SkewGP) (recently introduced in \cite{Benavoli_etal2020} for binary classification).\footnote{In particular, the present works extends the results in \cite{Benavoli_etal2020} by showing that SkewGPs are conjugate to \textit{probit affine likelihoods}. This allows us to apply SkewGPs for preference learning.}
Through several examples, we show  that the posterior has a
strong skewness, and thus  any approximation of the  posterior that relies on a symmetric distribution (such as Laplace's approximation)  results in sub-optimal predictive performances and, thus, slower convergence in PBO.

Second, we propose computationally efficient  methods to draw samples from the posterior that are then used to calculate the acquisition function.

Third, we extend standard acquisition functions used in BO to deal with preference observations and propose a new acquisition function for PBO obtained by combining the \textit{dueling information gain} with the expected  probability of improvement.

Fourth, we define an \emph{affine probit likelihood} to model the observations. Such a likelihood allows us to handle, in a unified framework, mixed categorical and preference observations. These two different  types of information  are usually available in manufacturing, where some  parameters may cause the process to fail and, therefore, producing no output. In standard BO, where the function evaluation is a scalar, a common way to address this problem is by penalizing the objective function when no output is produced, but this approach is not suitable in PBO  as the output is not a scalar.
 
 The rest of the paper is organized as follows. A review on skew-normal distributions and SkewGP is provided in Section~\ref{sec: Background}. The main results of the paper are reported in Section~\ref{sec:theory}, where we show that the posterior distribution of the latent preference function is a  SkewGP under the proposed affine probit likelihood. The marginal likelihood is derived and maximized to chose the model's hyper-parameters. An illustrative example is presented in  \ref{sec:comparison_surrogated} to show the drawbacks of Laplace's approximation, thus highlighting the  benefits of computing the exact SkewGP posterior distribution.  PBO through SkewGP is discussed in Section~\ref{sec:acquisition}, where extensive tests with different acquisition functions are reported and clearly show that PBO based on SkewGP  consistently outperforms Laplace's approximation both in terms of convergence speed and computational time.

\section{Background on skew-normal distributions and  skew-Gaussian processes} \label{sec: Background}
In this section we provide  details on the skew-normal distribution.
The skew-normal   \citep{o1976bayes,azzalini2013skew} is a large class of probability distributions that generalize a normal by allowing for non-zero skewness. 
A univariate skew-normal distribution is defined by three parameters location $\xi \in \mathbb{R}$, scale $\sigma>0$ and skew parameter $\alpha \in \mathbb{R}$ and has the following \citep{o1976bayes} Probability Density Function (PDF)
$$
p(z)={\frac {2}{\sigma }}\phi \left({\frac {z-\xi }{\sigma }}\right)\Phi \left(\alpha \left({\frac {z-\xi }{\sigma }}\right)\right), \qquad z \in \mathbb{R}
$$
where $\phi$ and $\Phi$ are the PDF and Cumulative Distribution Function (CDF), respectively, of the standard univariate Normal distribution.
Over the years many generalisations of this distribution were proposed, in particular \cite{arellano2006unification} provided a
unification of those generalizations in a
single and tractable multivariate \textit{Unified  Skew-Normal} distribution. This distribution satisfies closure properties for marginals and conditionals
and allows more flexibility due the introduction of additional
parameters.

\subsection{Unified Skew-Normal distribution}

A vector $\bz \in \mathbb{R}^p$ is distributed as a multivariate
Unified Skew-Normal distribution with latent skewness dimension $s$, $ \bz \sim \text{SUN}_{p,s}(\bxi,\Omega,\Delta,\bgamma,\Gamma)$, if its probability density function \citep[Ch.7]{azzalini2013skew} is:
\begin{equation}
\label{eq:sun}
p(\bz) = \phi_p(\bz-\bxi;\Omega)\frac{\Phi_s\left(\bgamma+\Delta^T\bar{\Omega}^{-1}\Domega^{-1}(\bz-\bxi);\Gamma-\Delta^T\bar{\Omega}^{-1}\Delta\right)}{\Phi_s\left(\bgamma;\Gamma\right)} 
\end{equation}
where $\phi_p(\bz-\bxi;\Omega)$ represents the PDF of a multivariate Normal distribution with mean $\bxi \in \mathbb{R}^p$ and covariance $\Omega=\Domega\bar{\Omega} \Domega\in \mathbb{R}^{p\times p}$,
with $\bar{\Omega}$ being a correlation matrix and $\Domega$ a diagonal matrix  containing the square root of the diagonal elements in $\Omega$. The notation $\Phi_s(\bl[a];M)$ denotes the CDF of $N_s(0,M)$ evaluated at $\bl[a]\in \mathbb{R}^s$. 
The parameters $\bgamma \in \mathbb{R}^s, \Gamma \in \mathbb{R}^{s\times s},\Delta^{p \times s}$ of the SUN distribution are related to a latent variable that controls the skewness, in particular $\Delta$ is called Skewness matrix.

	\begin{figure}[htp]
	\centering
	\begin{tabular}{c @{\quad} c }
		\includegraphics[width=.48\linewidth]{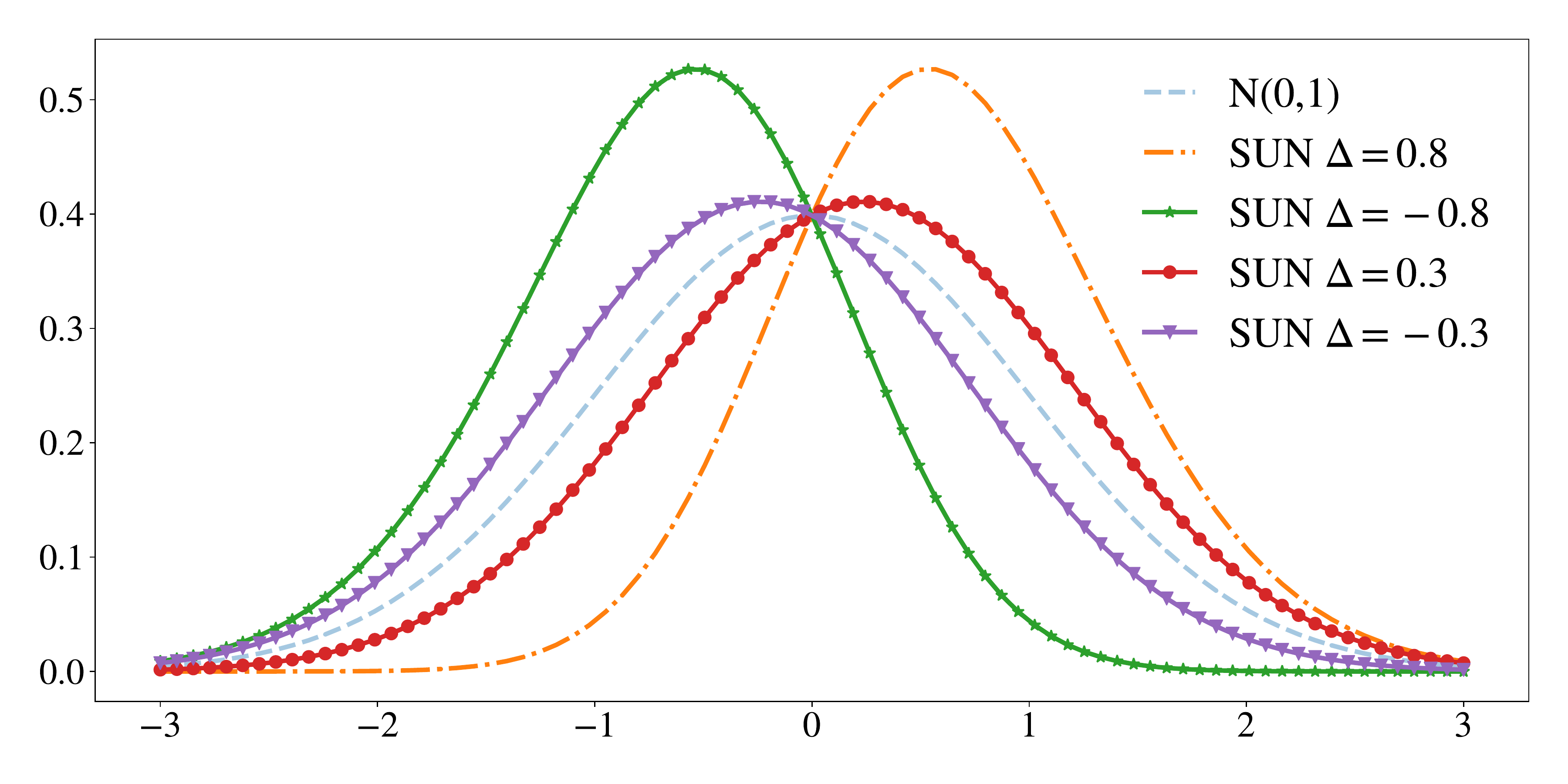} &
		\includegraphics[width=.48\linewidth]{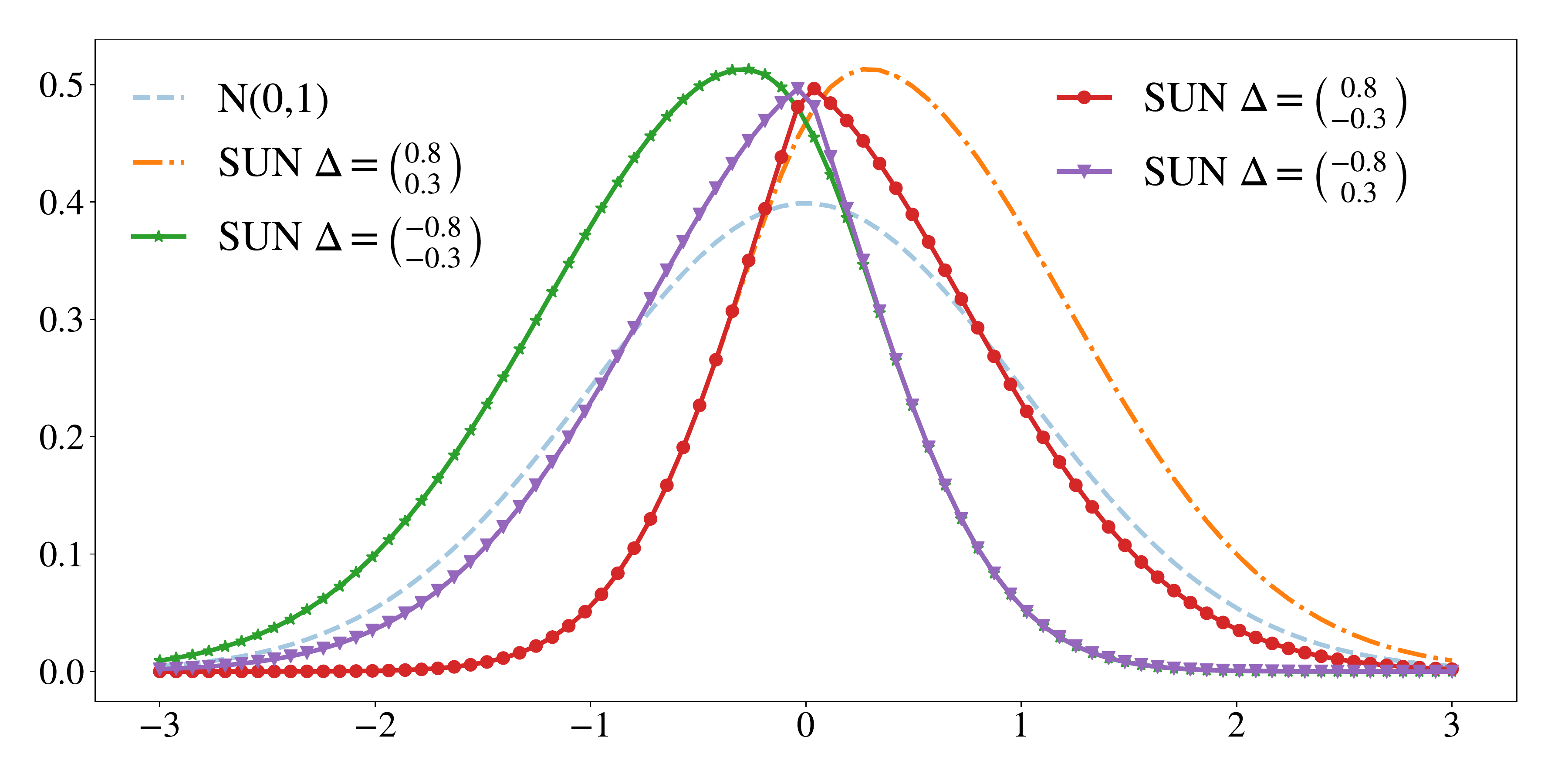} \\
		\small (a1) $s=1$, $\Gamma=1$  & \small (a2) $s=2$, $\Gamma_{1,2}=0.8$
	\end{tabular}
	\caption{Density plots for $\text{SUN}_{1,s}(0,1,\Delta,\gamma,\Gamma)$. For all plots $\Gamma$ is a correlation matrix, $\gamma = 0$, dashed lines are the contour plots of $y \sim N_1(0,1)$.}
	\label{fig:SUN1d}
\end{figure}

\begin{figure}
	\centering
	\begin{tabular}{c @{\quad} c @{\quad} c @{\quad} c}
		\includegraphics[width=.24\linewidth]{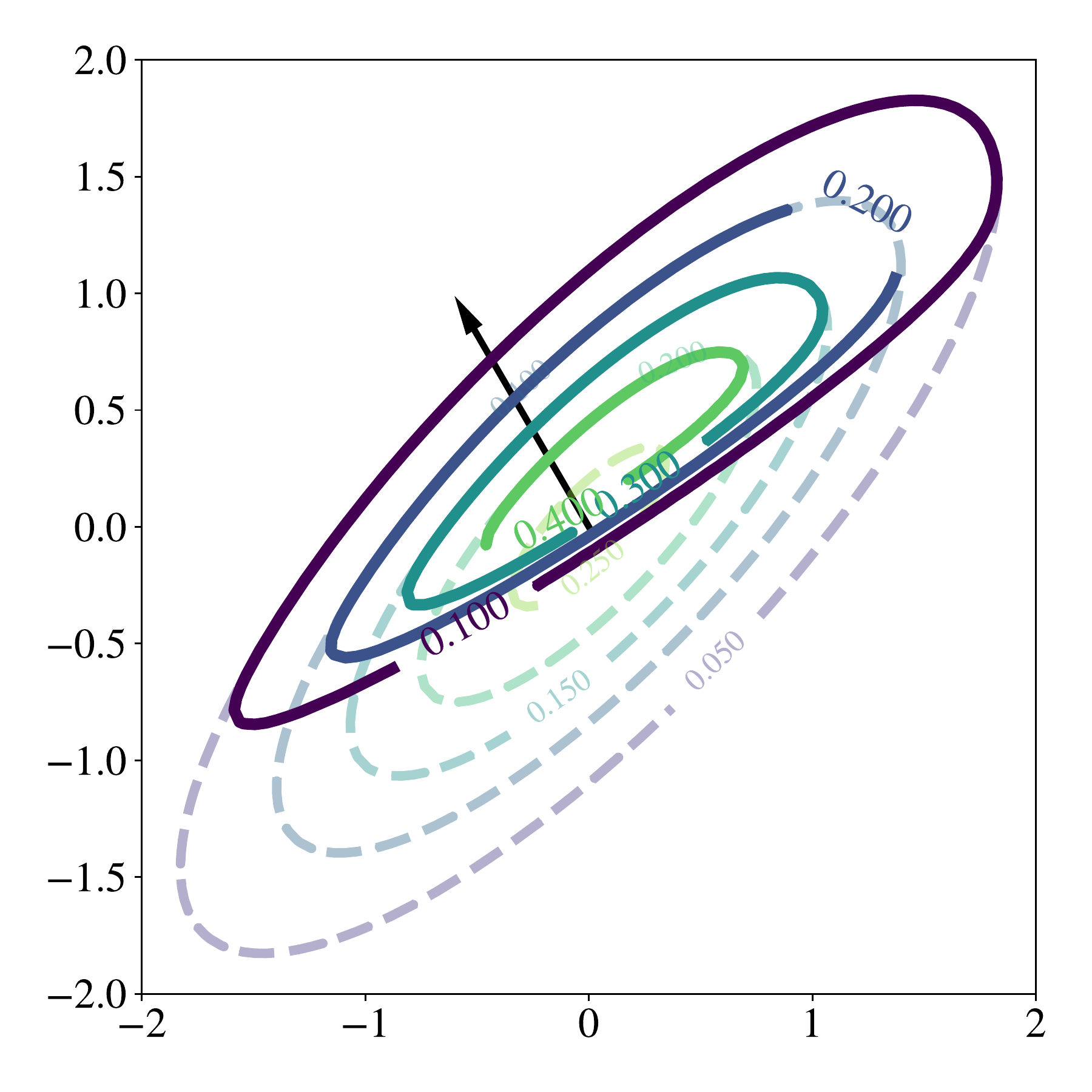} &
		\includegraphics[width=.24\linewidth]{{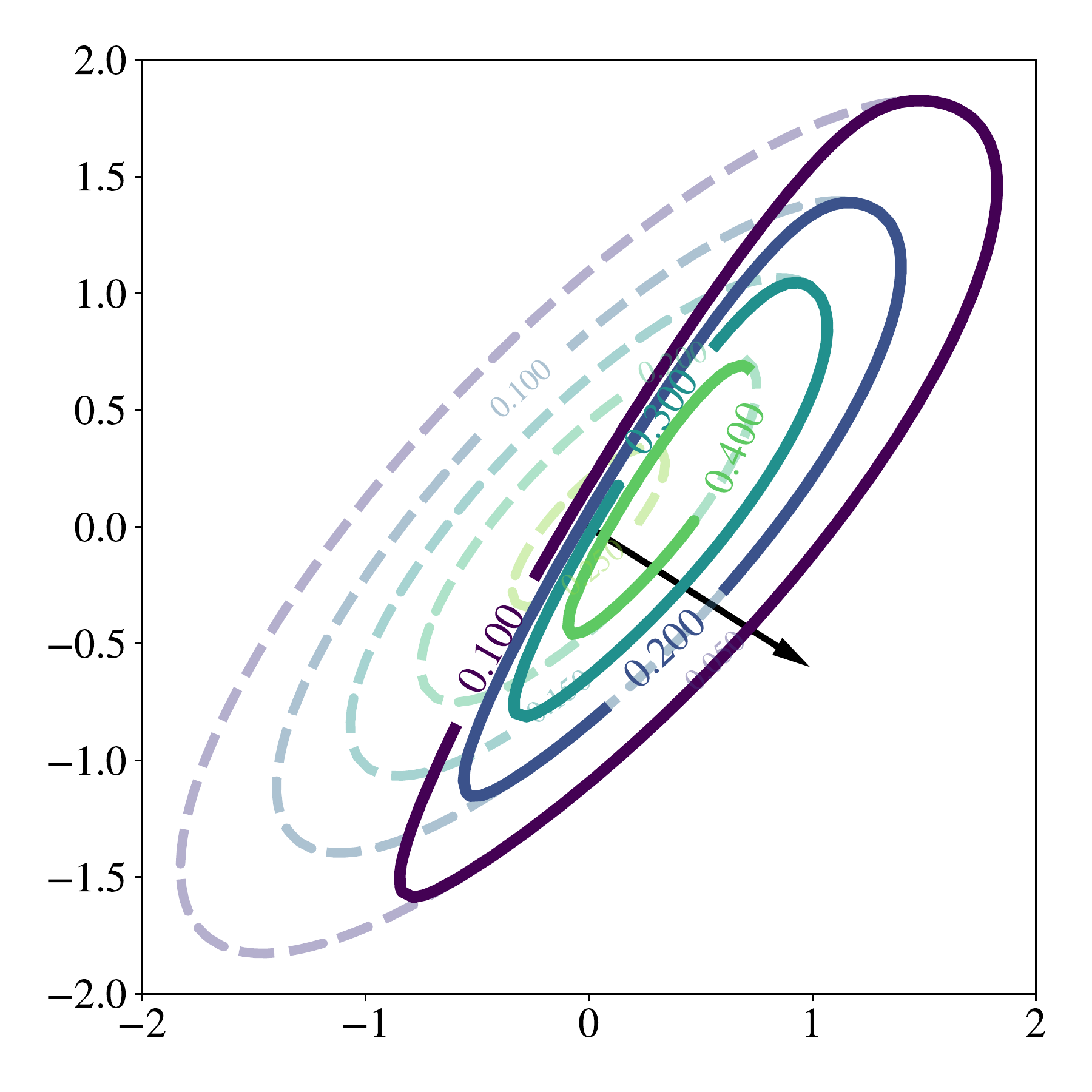}} &
		\includegraphics[width=.24\linewidth]{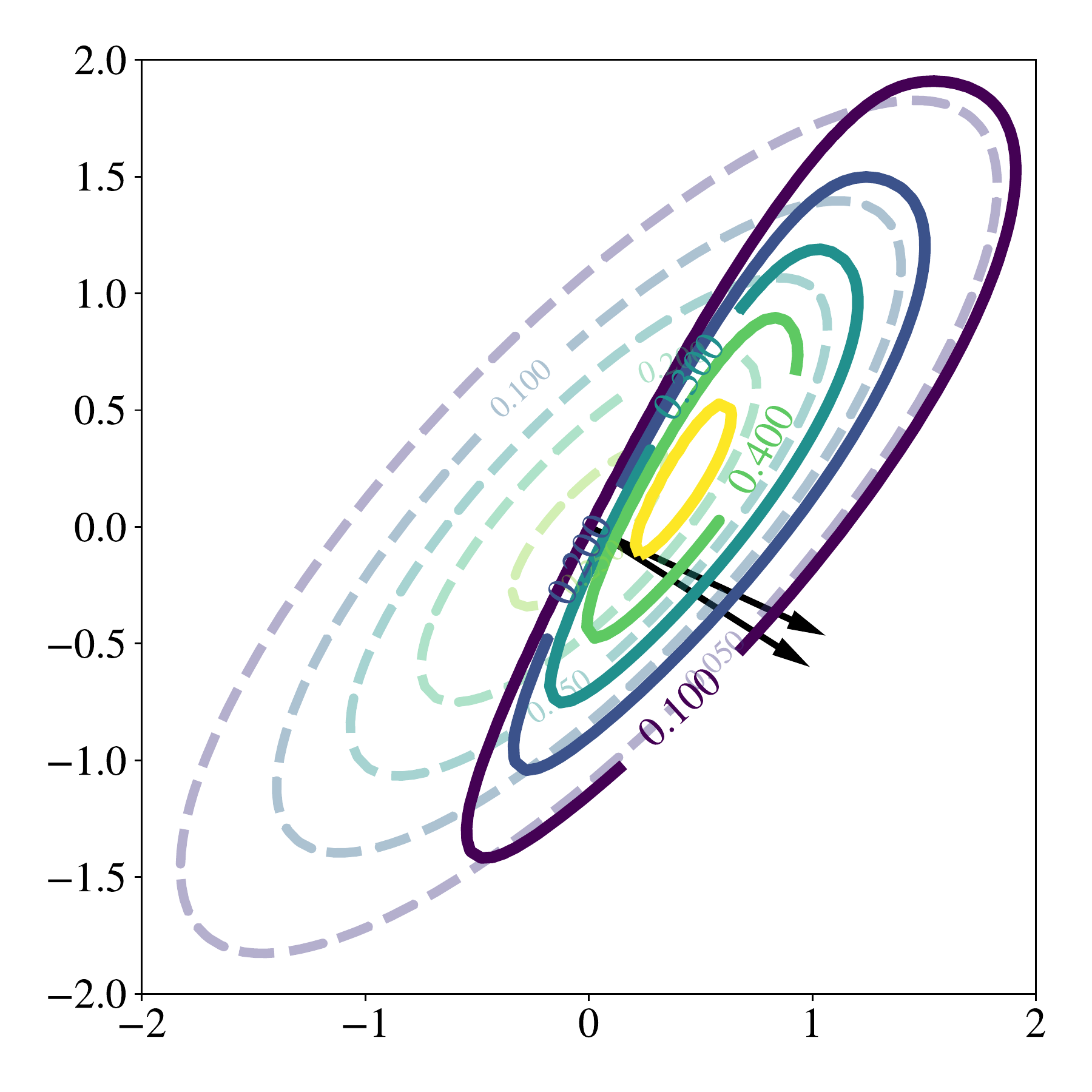} &
		\includegraphics[width=.24\linewidth]{{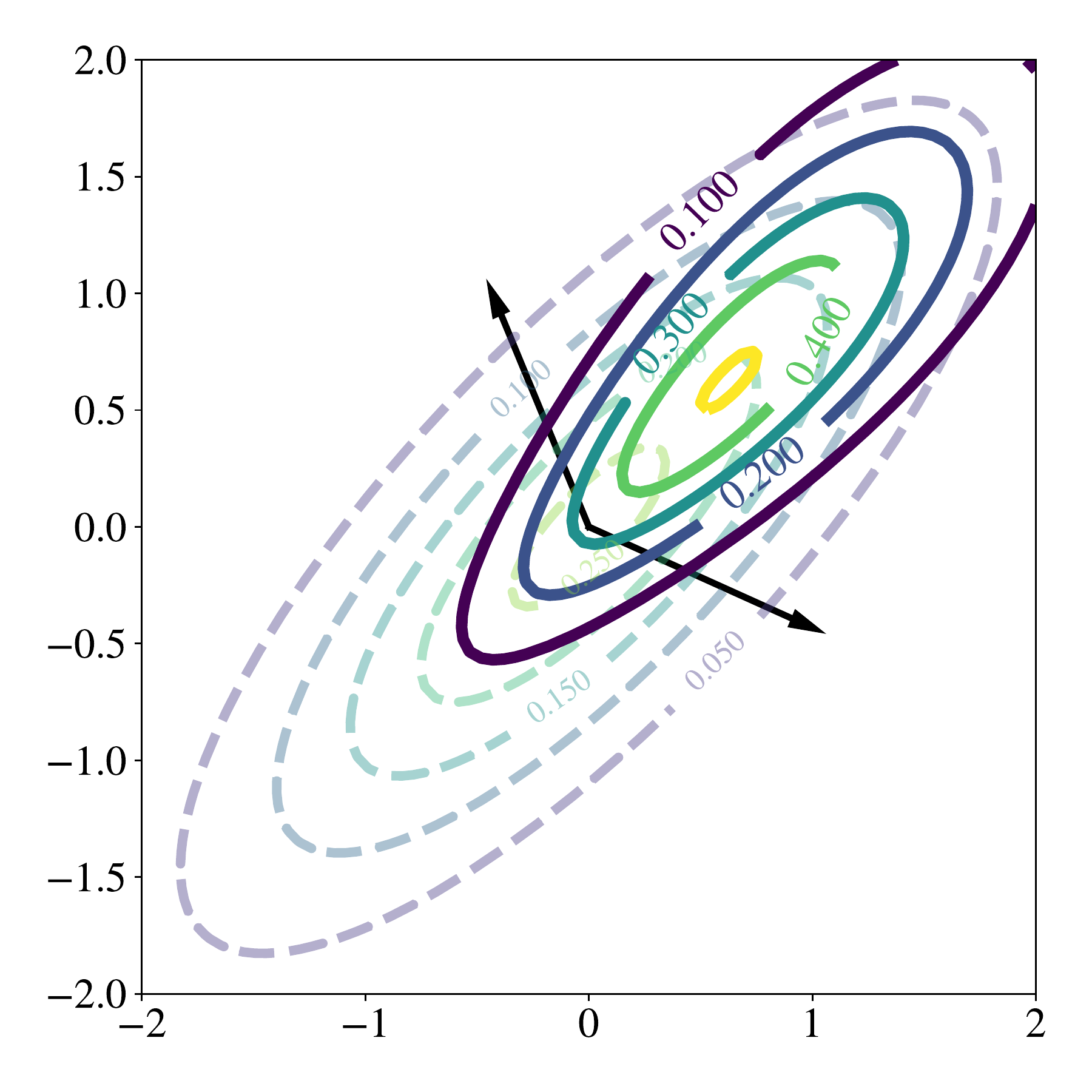}} \\
		\small (a1) $s=1$, $\Gamma=1$  & \small (a2) $s=1$, $\Gamma=1$ & \small (a3) $s=2$, $\Gamma_{1,2}=0.9$ & \small (a4) $s=2$, $\Gamma_{1,2}=-0.3$ \\
		\small $\Delta = [0.8,0.3]^T$,  & \small  $\Delta = [0.3,0.8]^T$ & \small $\Delta = \begin{bmatrix}
		0.3 & 0.5 \\
		0.8 & 0.9
		\end{bmatrix}$ & $\Delta = \begin{bmatrix}
		0.3 & 0.8 \\
		0.8 & 0.3
		\end{bmatrix}$ 
	\end{tabular}
	\caption{Contour density plots for four unified skew-normal. For all plots $p=2$, $\bxi=[0,0]^T$, $\Omega$ and $\Gamma$ are correlation matrices with $\Omega_{1,2}= 0.8$, $\gamma = 0$, dashed lines are the contour plots of $y \sim N_2(\bxi,\Omega)$.}
	\label{fig:SUN}
\end{figure}

The PDF  \eqref{eq:sun} is well-defined provided that the matrix
\begin{equation}
\label{eq:positivity}
M:=\begin{bmatrix}
\Gamma & \Delta^T\\
\Delta & \bar{\Omega}
\end{bmatrix} \in \mathbb{R}^{(s+p)\times(s+p)}>0,
\end{equation}
i.e., $M$ is positive definite. 
Note that when $\Delta=0$, \eqref{eq:sun}  reduces  to  $\phi_p(\bz-\bxi;\Omega)$, i.e. a skew-normal with zero skewness matrix is a normal distribution. Moreover we assume that $\Phi_0(\cdot)=1$, so that, for $s=0$, \eqref{eq:sun} becomes a multivariate Normal distribution.

Figure~\ref{fig:SUN1d} shows the density of a univariate SUN distribution with latent dimensions $s=1$ (a1) and $s=2$ (a2). The effect of a higher latent dimension can be better observed in bivariate SUN densities as shown in  Figure~\ref{fig:SUN}. 
The contours of the corresponding bivariate normal are dashed. We also plot the skewness directions given by $\bar{\Omega}^{-1}\Delta$.

For what follows however it is important to know that \citep[see, e.g.,][Ch.7]{azzalini2013skew} the distribution is closed under marginalization and conditioning. 
We have reviewed these results in  Appendix \ref{app:Background}  together with an \textit{additive representation} of the SUN that is useful for sampling from the posterior.


A SkewGP \cite{Benavoli_etal2020} is a generalization of a skew-normal distribution to a stochastic process. Its construction is based on a result derived by 
\cite{durante2018conjugate} for the parametric case, who showed that the skew-normal distribution and probit likelihood are conjugate.

To define a SkewGP, we consider here a location function $\xi: \mathbb{R}^d \rightarrow \mathbb{R}$, a scale (kernel) function $\Omega: \mathbb{R}^d \times \mathbb{R}^d \rightarrow \mathbb{R}$, a skewness vector function $\Delta: \mathbb{R}^d \rightarrow \mathbb{R}^s$ and the parameters $\bgamma \in \mathbb{R}^s, \Gamma \in \mathbb{R}^{s \times s}$. 
A real function $f: \mathbb{R}^d \rightarrow \mathbb{R}$ is a SkewGP with latent dimension $s$, if for any sequence of $n$ points $\bx_1, \ldots, \bx_n \in \mathbb{R}^d$, the vector $[f(\bx_1), \ldots, f(\bx_n)] \in \mathbb{R}^n$ is skew-normal distributed with parameters $\bgamma, \Gamma$ and location, scale and skewness matrices,
respectively, given by
\begin{equation}
\begin{array}{rl}
\xi(X):=\begin{bmatrix}
\xi(\bx_1)\\
\xi(\bx_2)\\
\vdots\\
\xi(\bx_n)\\
\end{bmatrix},~~
\Omega(X,X)&:=
\begin{bmatrix}
\Omega(\bx_1,\bx_1) & \Omega(\bx_1,\bx_2) &\dots & \Omega(\bx_1,\bx_n)\\
\Omega(\bx_2,\bx_1) & \Omega(\bx_2,\bx_2) &\dots & \Omega(\bx_2,\bx_n)\\
\vdots & \vdots &\dots & \vdots\\
\Omega(\bx_n,\bx_1) & \Omega(\bx_n,\bx_2) &\dots & \Omega(\bx_n,\bx_n)\\
\end{bmatrix},\vspace{0.2cm}\\
\Delta(X)&:=\begin{bmatrix}
~~\Delta(\bx_1) & \quad\quad\Delta(\bx_2) &~~\dots & ~\quad\Delta(\bx_n)\\
\end{bmatrix}.
\end{array}
\end{equation}

The skew-normal distribution is well defined if the matrix 
$M=\left[\begin{smallmatrix}
\Gamma & \Delta(X) \\
\Delta(X)^T & \Omega(X,X)
\end{smallmatrix}\right]
$
is positive definite for all $X = \{\bx_1, \ldots, \bx_n \} \subset \mathbb{R}^d$ and for any $n$. In that case we write $f \sim \text{SkewGP}_s(\xi,\Omega,\Delta,\gamma,\Gamma)$.
\citet{Benavoli_etal2020} shows that this is a well defined stochastic process. In the next section we connect 
  this stochastic process  to preference learning.

\section{SkewGP and affine probit likelihood}
\label{sec:theory}

Consider $n$ input points $X = \{\bx_i : i=1, \ldots, n\}$, with   $\bx_i \in \mathbb{R}^d$, and a data-dependent matrix $W \in \mathbb{R}^{m\times n}$. We define an affine probit likelihood as 
\begin{equation}
p(W \mid f(X)) = \Phi_m(Wf(X)),
\label{eq:affineProbit}
\end{equation}
where $\Phi_m(\bx) := \Phi_m(\bx; I_m)$ is the standard $m$-variate Gaussian CDF evaluated at $\bx \in\mathbb{R}^m$ with identity covariance matrix. 
Note that this likelihood model includes the classic GP probit classification model \citep{rasmussen2006gaussian} with binary observations $y_1, \ldots, y_n \in \{0,1\}$ encoded in the matrix
$
W=\text{diag}(2y_1-1, \ldots, 2y_n-1)
$, 
where $m=n$.
Moreover, as we will show in Corollary~\ref{cor:Wpref}, the likelihood in \eqref{eq:affineProbit} is equal to the preference likelihood for a particular choice of $W$. This model however also allows to seamlessly mix classification and preference information as we will show below. Here we prove how the skew-normal distribution is connected to the affine probit likelihood, extending a 
 a result proved in \citep[Th.1 and Co.4]{durante2018conjugate} for the parametric setting for a standard probit likelihood.\footnote{\citep[Th.1 and Co.4]{durante2018conjugate} assumes that the matrix $W$ is diagonal, but the same results can straightforwardly extend to generic $W$.}

\begin{theorem}
	\label{lemma:1}
	Let us assume that $f({\bf x})$ is GP distributed with mean function $\xi({\bf x})$ and covariance function $\Omega({\bf x},{\bf x}')$, that is
	$f({\bf x}) \sim \text{GP}(\xi({\bf x}), \Omega({\bf x},{\bf x}'))$, and consider the likelihood $p(W~\mid~f(X)) = \Phi_m(W f(X)) $ where $W\in \mathbb{R}^{m \times n}$.
	The posterior distribution of $f(X)$ is a SUN:
	\begin{align}
	\nonumber
	&p(f(X)|W)= \text{SUN}_{n,m}(\tilde{\xi},\tilde{\Omega},\tilde{\Delta},\tilde{\gamma},\tilde{\Gamma})~~~\text{ with }\\
		\label{eq:posteriorclass}
	&\tilde{\xi}  =\bxi, ~~~~ \tilde{\Omega} = \Omega, ~~~~
	\tilde{\Delta} =\bar{\Omega}\Domega W^T,~~~~
	\tilde{\gamma} =W\xi,~~~~
	\tilde{\Gamma}=W \Omega W^T + I_m,
	\end{align}
	where, for simplicity of notation, we denoted $\xi(X),\Omega(X,X)$ as $\bxi,\Omega$  and
	$\Omega = \Domega \bar{\Omega} \Domega$.
\end{theorem}
All the proofs are in  Appendix \ref{app:proofs}. We now prove that,  a-posteriori, for a new test point $\bf x$, the function  $f(\bf x)$ is SkewGP distributed under the affine probit likelihood in \eqref{eq:affineProbit}.

\begin{theorem}
	\label{th:1}
	Let us assume a GP prior 	$f({\bf x}) \sim \text{GP}(\xi({\bf x}), \Omega({\bf x},{\bf x}'))$, the likelihood $p(W \mid f(X)) = \Phi_m(W f(X)) $ with $W\in \mathbb{R}^{m \times n}$, then a-posteriori $f$ is SkewGP with mean function 
	$\xi({\bf x})$, covariance function $\Omega({\bf x},{\bf x}')$, skewness function $\Delta({\bf x},X)=\Omega({\bf x},X)W^T$, and 
	$\tilde{\gamma},\tilde{\Gamma}$ as in \eqref{eq:posteriorclass}.
\end{theorem}
This is the main result of the paper and  allows us to show  that, in the case of preference learning, we can compute exactly the posterior 
and, therefore, Laplace's approximation is not necessary.

\subsection{Exact preference learning}

We now apply results of Theorem~\ref{lemma:1} and Theorem~\ref{th:1} to the case of  preference learning. 
 For two different inputs $\bv_k , \bu_k \in X$, a pairwise preference $\bv_k \succ \bu_k$ is observed, where $\bv_k \succ \bu_k$ expresses the preference of the instance $\bv_k$ over $\bu_k$. A set of $m$  pairwise preferences is given and denoted as $\mathcal{D} =\{ \bv_k \succ \bu_k : k=1, \ldots, m \}$.

\paragraph{Likelihood.} We assume that there is an underlying hidden function $f: \mathbb{R}^d \rightarrow \mathbb{R}$ which is able to describe the observed set of pairwise preferences $\mathcal{D}$. Specifically, given a preference $\bv_k \succ \bu_k$, then the function $f$ is such that $f(\bv_k) \geq  f(\bu_k)$. To allow tolerances to model noise, we assume that value of the hidden function $f$ is corrupted by a Gaussian  noise with zero mean and  variance $\sigma^2$.

We use the likelihood introduced in \cite{ChuGhahramani_preference2005}, which is the joint probability distribution of observing the preferences $\mathcal{D}$ given the values of the function $f$ at $X$, i.e., 
\begin{align}
p(\mathcal{D} \mid f(X)) = &  \prod_{k=1}^{m} p(\bv_k \succ \bu_k|f(\bv_k), f(\bu_k)) =  \prod_{k=1}^{m} p(f(\bv_k)-f(\bu_k)\geq 0) = \nonumber \\
= & \prod_{k=1}^{m} \Phi\left(\frac{f(\bv_k)-f(\bu_k)}{\sqrt{2}\sigma}\right) = \Phi_m \left( \left(\begin{smallmatrix}
\frac{f(\bv_1)-f(\bu_1)}{\sqrt{2}\sigma}\ \\
\vdots \\
\frac{f(\bv_m)-f(\bu_m)}{\sqrt{2}\sigma}
\end{smallmatrix}\right) \right). \label{eqn:likel}
\end{align}
For identifiability reasons, without loss of generality, we set $\sigma^2= \frac{1}{2}$.\footnote{Equivalently, we instead estimate the kernel variance.} 

\paragraph{Posterior.}

The posterior distribution of the values of the hidden function $f$ at all $\bx \in X$ given the observations $\mathcal{D}$ is then:
\begin{align}
p( f(X) \mid \mathcal{D}) = \frac{p(f(X))}{p( \mathcal{D})}   \Phi_m \left( \begin{pmatrix}
 f(\bv_1)-f(\bu_1) \ \\
\cdots \\
 f(\bv_m)-f(\bu_m) \
\end{pmatrix}\right).
\end{align}

In state-of-art PBO \citep{gonzalez2017preferential}, a Laplace's  approximation of the posterior  $p( f(X) \mid \mathcal{D})$   is used to  construct the acquisition function. The following Corollary shows that the posterior $p( f(X) \mid \mathcal{D})$ is distributed as a  SkewGP. 


\begin{corollary}
	\label{cor:Wpref}
	Consider  $f({\bf x}) \sim \text{GP}(\xi({\bf x}), \Omega({\bf x},{\bf x}'))$ and the likelihood  $p(\mathcal{D} \mid f(X))$ in \eqref{eqn:likel}. If we denote by $W \in \mathbb{R}^{m \times n}$ the matrix defined as $W_{i,j} = V_{i,j} - U_{i,j}$ where $V_{i,j}=1$ if $\bv_i=\bx_j$ and $0$ otherwise and $U_{i,j}=1$ if $\bu_i=\bx_j$ and $0$ otherwise. Then the posterior of $f(X)$ is given by \eqref{eq:posteriorclass}. 
\end{corollary}

In PBO, in order to compute the acquisition functions, we must be able to draw efficiently independent samples from the posterior in Theorem \ref{th:1}.
\begin{proposition}
	\label{po:1}
Given a   test point ${\bf x}$, posterior samples of $f({\bf x})$ can be obtained as:
\begin{align}
\label{eq:sampling}
f({\bf x}) &\sim \tilde{\xi}({\bf x})+ D_{\Omega({\bf x},{\bf x})}\left(U_0+\Omega({\bf x},X)W^T(W \Omega(X,X) W^T + I_m)^{-1}U_1\right),\\
\nonumber
U_0 &\sim \mathcal{N}(0;\bar{\Omega}({\bf x},{\bf x})-\Omega({\bf x},X)W^T\tilde{\Gamma}^{-1}W\Omega({\bf x},X)^T),\qquad 
U_1 \sim \mathcal{T}_{\tilde{\bgamma}}(0;W \Omega(X,X) W^T + I_m), 
\end{align}
where $\mathcal{T}_{\tilde{\bgamma}}(0;\tilde{\Gamma})$ is the pdf of a multivariate Gaussian distribution with zero mean and covariance $\tilde{\Gamma}$  truncated component-wise below $-\tilde{\bgamma}=-W\xi(X)$.
\end{proposition}
 Note that sampling $U_0$ can be achieved efficiently with standard methods, however using standard rejection sampling for the variable $U_1$ would incur in exponentially growing sampling time as the dimension $m$ increases.
Here we use the recently introduced sampling technique \emph{linear elliptical slice sampling} (\emph{lin-ess},\citet{gessner2019integrals}) which improves Elliptical Slice Sampling (\emph{ess}, \citet{pmlrv9murray10a}) for multivariate Gaussian distributions truncated on a region defined by linear constraints. In particular this approach derives analytically the acceptable regions on the elliptical slices used in \emph{ess} and guarantees rejection-free sampling.
Since \emph{lin-ess} is rejection-free,\footnote{Its computational bottleneck is 
 the Cholesky factorization of the covariance matrix $\tilde{\Gamma}$, same as for sampling from a multivariate Gaussian.} we can compute exactly  the computation complexity of \eqref{eq:sampling}: $O(n^3)$ with storage demands of $O(n^2)$. SkewGPs have similar bottleneck computational complexity of full GPs.
Finally, observe that  $U_1$ does not depend on ${\bf x}$
and, therefore, we do not need to re-sample $U_1$ to sample
$f$ at another test point ${\bf x}'$. This is fundamental  because   acquisition functions are functions of ${\bf x}$ and, we need to optimize them in PBO.


\paragraph{Marginal likelihood.}  Here, we follow the usual GP literature (\cite{rasmussen2006gaussian}) and we consider a zero mean function $\xi({\bf x})=0$ and a parametric covariance kernel $\Omega({\bf x},{\bf x}')$ indexed by $\theta \in \Theta$.Typically $\theta$ contains lengthscale parameters and a variance parameter.
For instance, for the RBF kernel
 $$
\Omega({\bf x},{\bf x}') := \sigma^2 \exp \left(-\frac {\|{\bx} -{\bx'} \|^{2}}{2\ell^{2}}\right).
 $$
 we have that $\theta=[\ell,\sigma]$.\footnote{In the numerical experiments, we use a RBF kernel with ARD and so we have a lengthscale parameter for each component of ${\bf x}$.}   The parameters $\theta$ are chosen by maximizing the marginal likelihood, that for SkewGP is provided hereafter.
\begin{corollary}
	\label{co:ml}
	Consider a GP prior $f({\bf x}) \sim \text{GP}(\xi({\bf x}), \Omega({\bf x},{\bf x}'))$ and the likelihood $p(\mathcal{D} \mid f(X)) = \Phi_m(Wf(X))$, then the marginal likelihood of the observations $\mathcal{D}$ is
	\begin{equation}
	p(\mathcal{D}) =  \Phi_{m}(\tilde{\bgamma};~\tilde{\Gamma}) ~~~\left(\geq  \sum_{i=1}^{b} \Phi_{|B_i|}(\tilde{\bgamma}_{B_i};~\tilde{\Gamma}_{B_i})-(b-1)\right),
	\label{eq:marginalLikelihood}
	\end{equation}
	with $\tilde{\bgamma},\tilde{\Gamma}$ defined in Theorem
	\ref{th:1} (they depend on $\theta$).
\end{corollary}
If the size of $W$ is too large the evaluation of $\Phi_{m}$ could become infeasible, therefore here we use the  approximation introduced in \cite{Benavoli_etal2020}, 
see inequality in \eqref{eq:marginalLikelihood} where $B_1,\dots,B_b$ are a partition of the training dataset  into $b$ random disjoint subsets, $|B_i|$ denotes the number of observations  in the i-th element of the partition, 
$\tilde{\bgamma}_{B_i},~\tilde{\Gamma}_{B_i}$ are the parameters
of the posterior computed using only the subset $B_i$ of the data (in the experiments $|B_i|=30$). Details about the  routine we use to compute $\Phi_{|B_i|}(\cdot)$ and the optimization method we employ to maximise the lower bound  in \eqref{eq:marginalLikelihood} are  in  Appendix \ref{app:implementation}.

\subsection{Mixed classification and preference information}
\label{subsec:mixed}
	
Consider now a problem where we have two types of information: whether a certain instance is preferable to another (preference-like observation) and whether a certain instance is attainable or not (classification-like observation). Such situation often comes up in industrial applications. For example imagine a machine that produces a product whose final quality depends on certain input parameters. Assume now that certain values of the input parameters produce no product. In this case we might want to evaluate the quality of the product with binary comparisons (preferences) along with a binary class that indicates whether the input configuration is valid. By using only a preference likelihood or a classification likelihood we would not be using all information. 
In this case,  observations are in fact  pairs and the space of possibility is $
\mathcal{Z}=\{(\text{valid},{\bv_k}\succ {\bu_k}),~~(\text{valid},{\bu_k}\succ {\bv_k}),~~(\text{non-valid},None)\},
$
where ${\bv_k}$ and ${\bu_k}$ are respectively the current and reference input.
We propose a new likelihood function  to model the above setting, which is defined as follows:
\begin{equation*}
P(z_k|f({\bv_k}),f({\bu_k}))=\left\{\begin{array}{ll}
\Phi\left(f({\bv_k})\right)\Phi\left(f({\bv_k})-f({\bu_k})\right), & z_k=   (\text{valid},{\bv_k}\succ {\bu_k})\\                    
\Phi\left(f({\bv_k})\right)\Phi\left(f({\bu_k})-f({\bv_k})\right), & z_k=   (\text{valid},{\bu_k}\succ {\bv_k})\\
\Phi\left(-f({\bv_k})\right), & z_k=   (\text{non-valid},None).\\
\end{array}\right.
\end{equation*}
It is then immediate to 
write the above likelihood 
in the form \eqref{eq:affineProbit} 
and, therefore, use both sources of information.
We associate to each point $\bx_i$ a binary output $y_i \in \{0,1\}$ where the class $0$ denotes a non-valid output. In case of valid output,  we assume that we conducted $m$ comparisons obtaining the couples $\mathcal{D} =\{ \bv_k \succ \bu_k : k=1, \ldots, m \}$ where $\bv_k \succ \bu_k$ expresses the preference of the instance $\bv_k$ over $\bu_k$. The likelihood is then a product of two independent probit likelihood functions 
\begin{align*}
\small
p_{\text{class}}(W_{\text{class}} \mid f(X)) &= \Phi_n\left(\left[\begin{smallmatrix}
2y_1 -1 & 0 & \cdots & 0 \\
0 & 2y_2 -1 & \cdots & 0 \\
\vdots & \vdots & \ddots & \vdots \\
0 & 0 & \cdots & 2y_n -1 \\
\end{smallmatrix}\right]f(X) \right), ~
p_{\text{pref}}(W_{\text{pref}} \mid f(X)) &= \Phi_m\left(\left[\begin{smallmatrix}
f(\bv_1) - f(\bu_1) \\
\vdots \\
f(\bv_m) - f(\bu_m)
\end{smallmatrix}\right] \right)
\end{align*}
Since we assume that the two likelihood are independent we can compute the overall likelihood: 
\begin{align} \label{eq:mixLikelihood}
p(W_{\text{class}},W_{\text{pref}}  \mid f(X)) &= p_{\text{class}}(W_{\text{class}} \mid f(X)) p_{\text{pref}}(W_{\text{pref}} \mid f(X)) = \Phi_{n+m} (Wf(X)), 
\end{align}
with $W=\begin{bmatrix}W_{\text{class}}\\W_{\text{pref}}\end{bmatrix}\in\mathbb{R}^{(n+m)\times n}$,
where $W_{\text{pref}}$ is the matrix of preferences defined as in Corollary~\ref{cor:Wpref}. Therefore, the results
in Section \ref{sec:theory} still holds in this mixed  setting.



%
 \section{Comparison SkewGP vs.\ Laplace's approximation}
 \label{sec:comparison_surrogated}

\begin{figure}
	\centering
	\begin{tabular}{ll}
		\begin{minipage}{7cm}
			\includegraphics[height=4.9cm,trim={0.0cm 0.0cm 0.0cm 0.0cm }, clip]{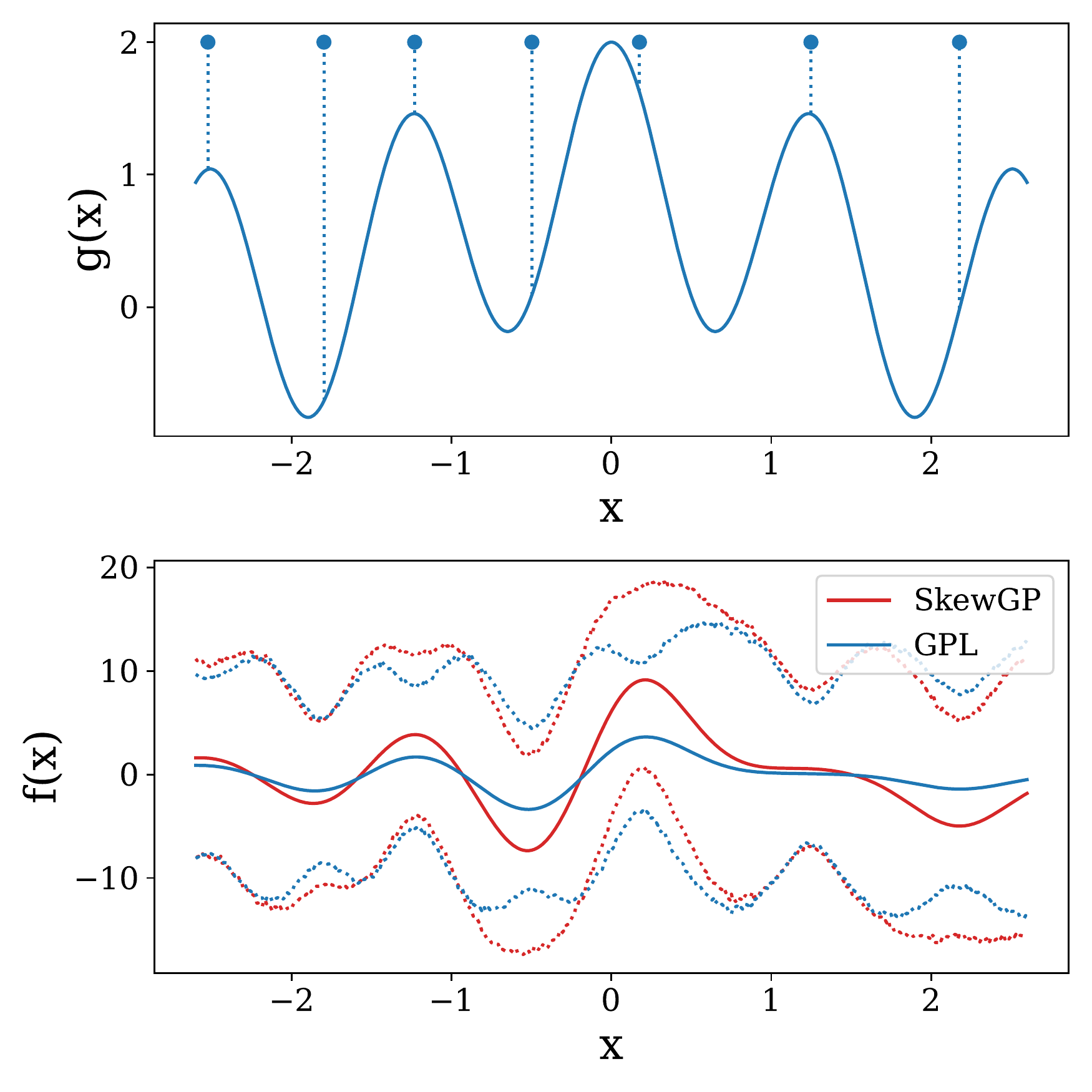}
		\end{minipage}
		\hspace{-1.4cm}\begin{minipage}{4.5cm}
			\includegraphics[height=2.4cm]{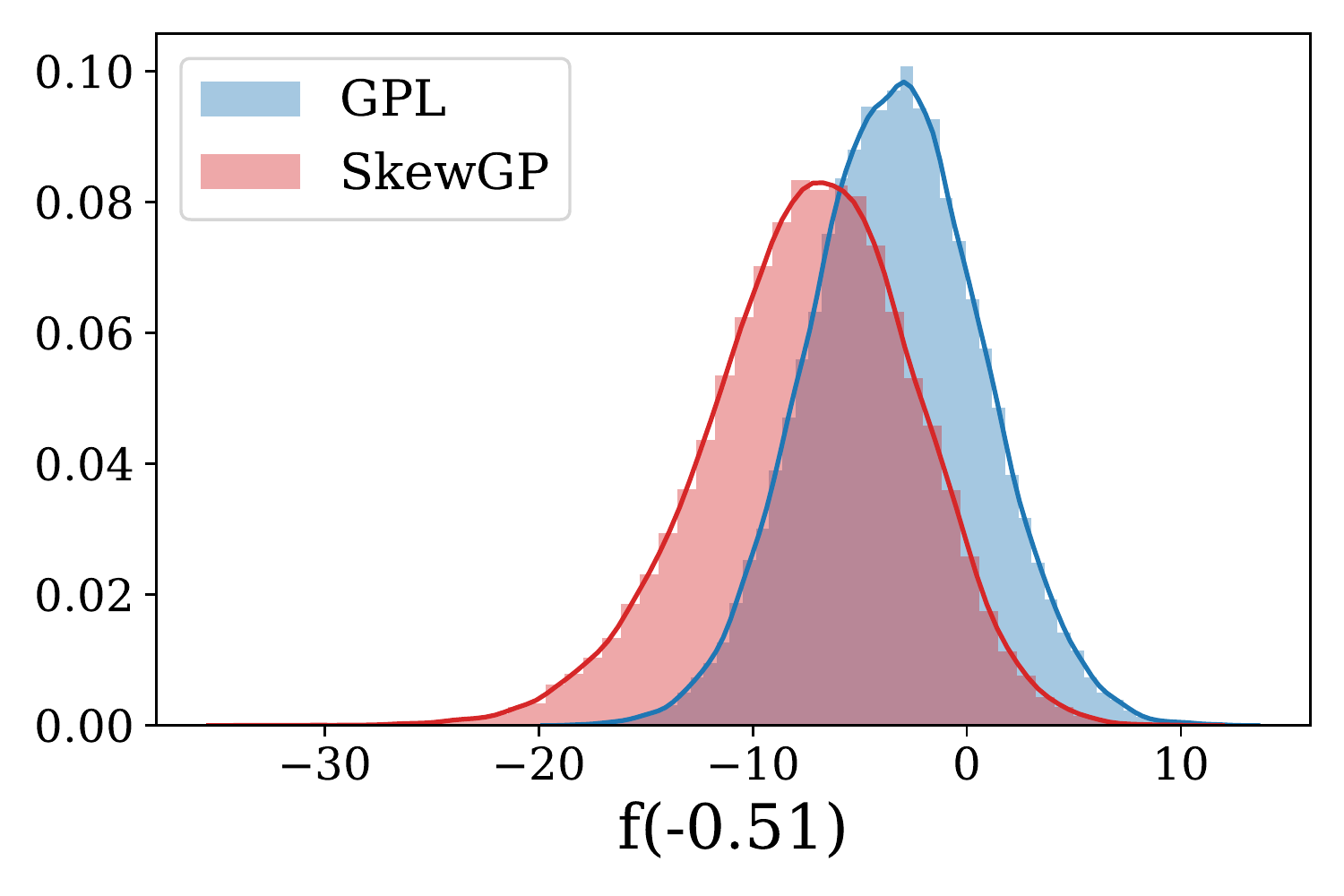}\\
			\includegraphics[height=2.4cm]{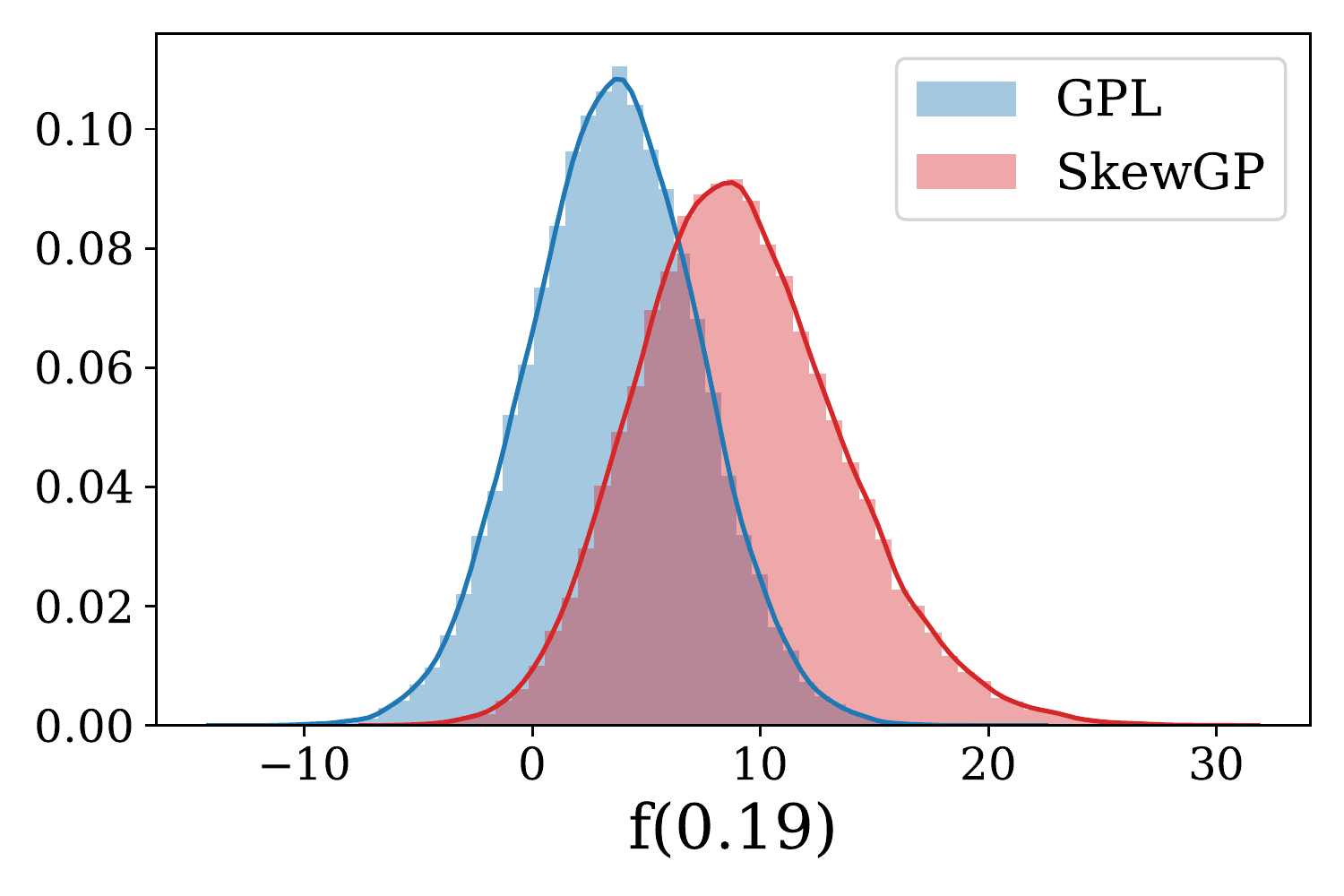}
		\end{minipage}
	\end{tabular}
	\caption{Comparison between Laplace's approximation (GPL)
		and the exact posterior (SkewGP). Top-left shows $g(x)$. Bottom-left shows the  predicted posterior preference function $f(x)$ (continuous lines) and the relative 95\% credible region (dashed lines) for GPL and SkewGP. Right-column reports the density plots for $f(0.19)$ (bottom) and 
		$f(-0.51)$ (top) for both models.}
	\label{fig:2}
\end{figure}

We  provide a one-dimensional illustration of the difference between GPL and SkewGP.
Consider the  non-linear function $g(x)=cos(5x)+e^{-\frac{x^2}{2}}$ which has a global maximum at $x=0$.
We assume we can only query this function through pairwise comparisons. We generate
7 pairwise random comparisons: the query point $x_i$ is preferred to $x_j$ (that is  $x_i\succ x_j$) if $g(x_i)>g(x_j)$.
Figure  \ref{fig:2}(top-left) shows $g(x)$ and the location of the queried points.\footnote{The preferences between the queried points are  $1.25\succ-1.8$,
	$-1.23\succ1.25$,
	$0.18\succ-1.23$,
	$0.18\succ-2.52$,
	$-2.52\succ2.18$,
	$-1.8\succ-0.5$,
	$-1.8\succ0.67$.}
Figure  \ref{fig:2}(bottom-left) shows the   predicted posterior preference function $f(x)$ (and relative 95\% credible region) computed according to GPL and SkewGP. Both the methods  have the same prior: a GP with zero mean and RBF covariance function (the hyperparameters are the same for both methods and have been set equal to the values that maximise  Laplace's approximation to the marginal likelihood, $l=0.35$ and $\sigma^2=0.02$).
Therefore, the only difference between the two posteriors
is due to the Laplace's approximation. The true posterior (SkewGP) of the preference function is skewed, this can be seen from the density plot for  $f(-0.51)$ in Figure \ref{fig:2}(top-right). 
Figure \ref{fig:2}(bottom-right) 
shows an example, $f(0.19)$,  where SkewGP and Laplace's approximation differ significantly:  Laplace's approximation 
heavily underestimates the mean  and   the support
of the true posterior (SkewGP) also evident from Figure \ref{fig:2}(bottom-left).
These differences determine the poor performance of PBO  based on GPL as we will see in the next sections.

\section{Dueling acquisition functions}
\label{sec:acquisition}

In sequential BO, our objective is to seek a new data point ${\bf x}$ which will allow us to
get closer to the maximum of the target function $g$.
Since $g$ can only be queried via preferences, this is obtained by optimizing w.r.t.\ ${\bf x}$ a dueling acquisition function $\alpha({\bf x},{\bf x}_r)$, where ${\bf x}_r$ (reference point) is the best point found so far, that is the point that has the highest probability of
winning most of the duels (given the observed data $\mathcal{D}$) and, therefore, it is the most likely point maximizing $g$.\footnote{ By optimizing the acquisition function $\alpha({\bf x},{\bf x}_r)$, we aim to find a point that is better than ${\bf x}_r$ (also considering the trade-off between exploration and exploitation). After computing the optimum of the the acquisition function, denoted with ${\bf x}_n$, we query the black-box function for ${\bf x}_n\stackrel{?}{\succ} {\bf x}_r$. If    ${\bf x}_n\succ {\bf x}_r$  then ${\bf x}_n$ becomes the new reference point (${\bf x}_r$) for the next iteration.}  
We consider three pairwise modifications of standard acquisition functions: (i) Upper Credible Bound (UCB); (ii)  Thompson sampling (TH); (iii) Expected Improvement Info Gain (EIIG). 

 {\bf UCB:} The dueling UCB acquisition function is defined as the upper bound of the minimum width $\gamma$\% (in the experiments we use $\gamma=95$) credible interval of $f({\bf x})-f({\bf x}_r)$.

 {\bf TH:} The dueling Thompson acquisition function is  
$
f_j({\bf x})-f_j({\bf x}_r), 
$
where $f_j$ is a sampled function from the posterior.

{\bf EIIG:} We now propose the dueling EIIG that is  the combination of the expected  probability of improvement (in log-scale) and the dueling information gain: 
$$
k \log\left(E_{f \sim p(f|\mathcal{D})}\left(\Phi\left(\tfrac{f({\bf x})-f({\bf x}_r)}{\sqrt{2}\sigma}\right)\right)\right) - IG({\bf x},{\bf x}_r),$$
where
$
IG({\bf x},{\bf x}_r)=h\left(E_{f \sim p(f|\mathcal{D})}\left(\Phi\left(\tfrac{f({\bf x})-f({\bf x}_r)}{\sqrt{2}\sigma}\right)\right)\right)-E_{f \sim p(f|\mathcal{D})}\left(h\left(\Phi\left(\tfrac{f({\bf x})-f({\bf x}_r)}{\sqrt{2}\sigma}\right)\right)\right),
$   
with 
$h(p)=-p \log(p)-(1-p)\log(1-p)$ being the binary entropy function of  $p$.
This definition of  dueling information gain is an extension  to preferences of the  information gain, formulated for GP classifiers in \citep{houlsby2011bayesian}.
This last acquisition function allows us to balance exploration-exploitation by means of the nonnegative scalar
$k$ (in the  experiments we use $k=0.1$ (more exploration) and $k=0.5$).
 To compute these acquisition functions, we make explicitly use of the generative capabilities of our SkewGP surrogated model as well as the efficiency
 of sampling the learned preference function.

 Note that, \cite{gonzalez2017preferential} use different acquisition functions based on the Copland's score (to increase exploration).
 Moreover, they optimize $\alpha({\bf x}_a,{\bf x}_b)$
 with respect to both ${\bf x}_a,{\bf x}_b$, while 
 ${\bf x}_b$ is fixed and equal to ${\bf x}_r$ in our setting.
  SkewGP can easily be employed as surrogated model in \cite{gonzalez2017preferential} PBO setting (and improve their performance due to the limits  of the Laplace's approximation). We have focused on the above 
  acquisition functions (UCB, TH, EIIG) because they can easily be computed -- instead the Copland's score requires to numerically compute an integral with respect to ${\bf x}$.

\section{Numerical experiments}
\label{sec:Numerical experiments}

In this section we present numerical experiments to validate
our  PBO-SkewGP and compare it with PBO based on the Laplace's approximation (PBO-GPL).

First, we consider again the maximization of $g(x)=cos(5x)+e^{-\frac{x^2}{2}}$ and the same $7$
initial preferences used in Section \ref{sec:comparison_surrogated}.
\begin{figure}
\centering
	\includegraphics[height=3cm,trim={0.0cm 0.0cm 0.0cm 0.0cm }, clip]{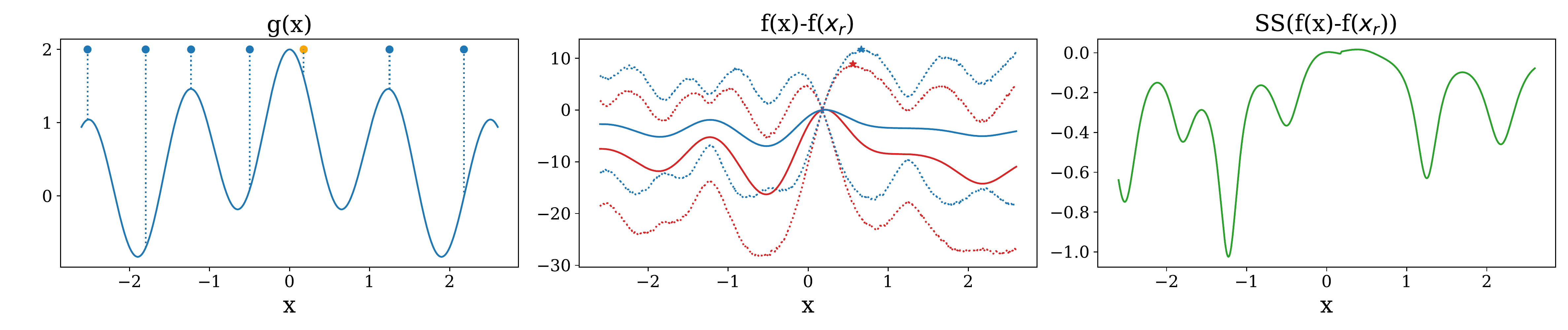}\\
	\includegraphics[height=3cm,trim={0.0cm 0.0cm 0.0cm 0.0cm }, clip]{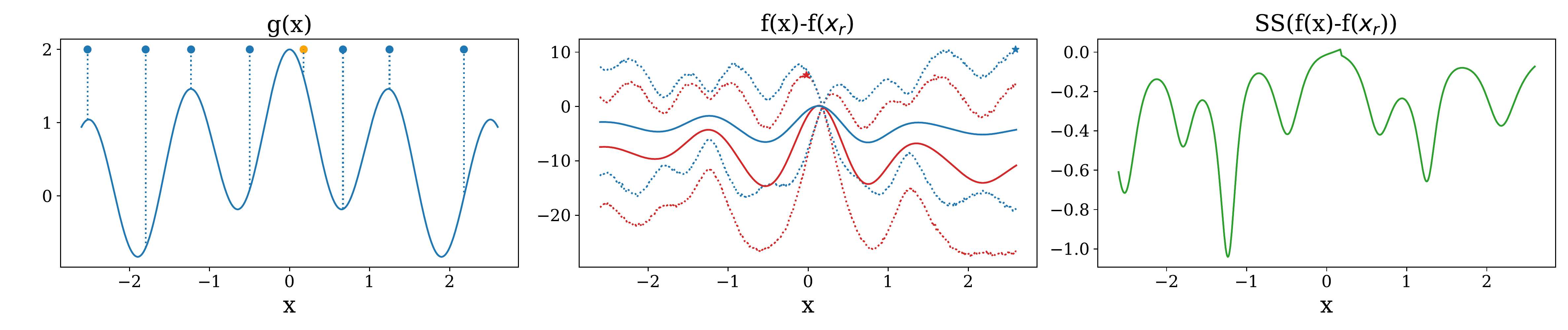}\\
	    \includegraphics[height=3.cm,trim={0.0cm 0.0cm 0.0cm 0 }, clip]{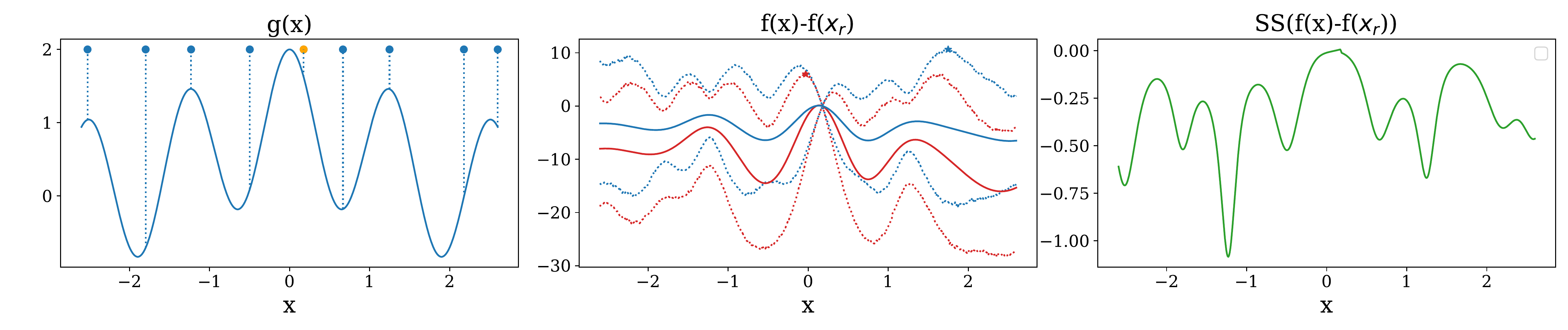}\\
	     \includegraphics[height=3.cm,trim={0.0cm 0.0cm 0.0cm 0 }, clip]{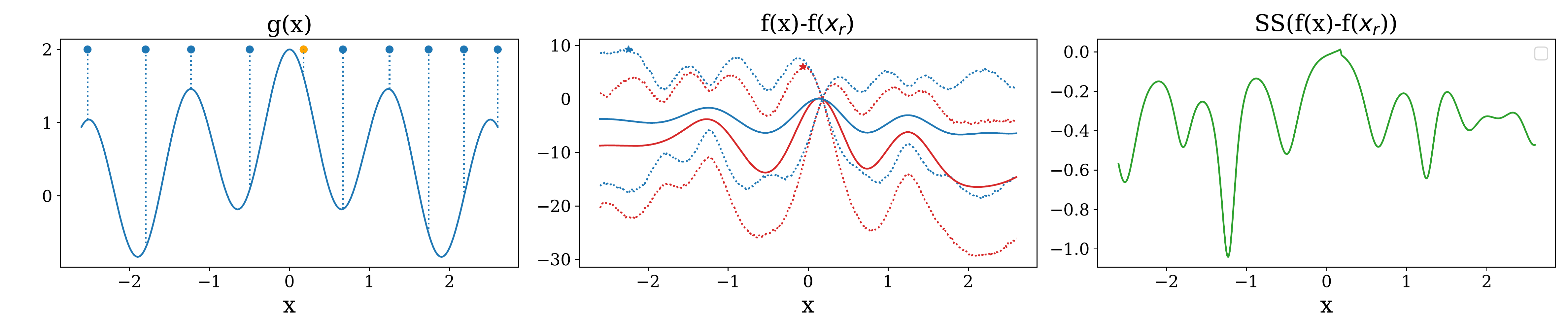}
	\caption{PBO run for 4 iterations. In each iteration, a row with three plots is showed. The left plot shows the  objective function and the queried points ($x_r$ is in orange). The central plot shows the GPL (blue) and SkewGP (red) posterior predictive means of $f(x)-f(x_r)$ and the relative  95\% credible intervals. The maximum of each UCB is showed with a star marker. The right plot shows the skewness statistics for the SkewGP predictive posterior distribution of $f(x)-f(x_r)$. At each step, we query the point that maximises the UCB of GPL.} \label{fig:3}
\end{figure}

We run PBO for 4 iterations and, at each step, we \textbf{query the point that maximises the UCB of GPL}. 
We also compute the true posterior and the true maximum of UCB using SkewGP for comparison.  Both the methods  have the same prior: a GP with zero mean and RBF covariance function (the \textbf{hyperparameters are fixed to the same values for both methods}, that is the values that maximise  Laplace's approximation to the marginal likelihood, $l=0.35$ and $\sigma^2=0.02$). Therefore, \textbf{the only difference between the two posteriors
is due to the Laplace's approximation}.

Figure \ref{fig:3} shows, for each iteration, a row with three plots. The left plot reports  $g(x)$ and the queried points ($x_r$ is in orange). The central plot shows the GPL and SkewGP posterior predictive means of $f(x)-f(x_r)$ and the relative  95\% credible intervals. The maximum of each UCB is showed with a star marker. The right plot shows the skewness statistics for the SkewGP predictive posterior distribution of $f(x)-f(x_r)$ as a function of $x$, defined as:
$$
SS(f(x)-f(x_r)):={\tfrac {\operatorname {E} \left[(f(x)-\mu )^{3}\right]}{(\operatorname {E} \left[(f(x)-\mu )^{2}\right])^{3/2}}},
$$
with $\mu:=\operatorname {E} \left[f(x)\right]$, and computed  via Monte Carlo sampling  from the posterior.

\begin{figure}
	\centering
	\begin{tabular}{cc}
 		\includegraphics[width=.3\linewidth]{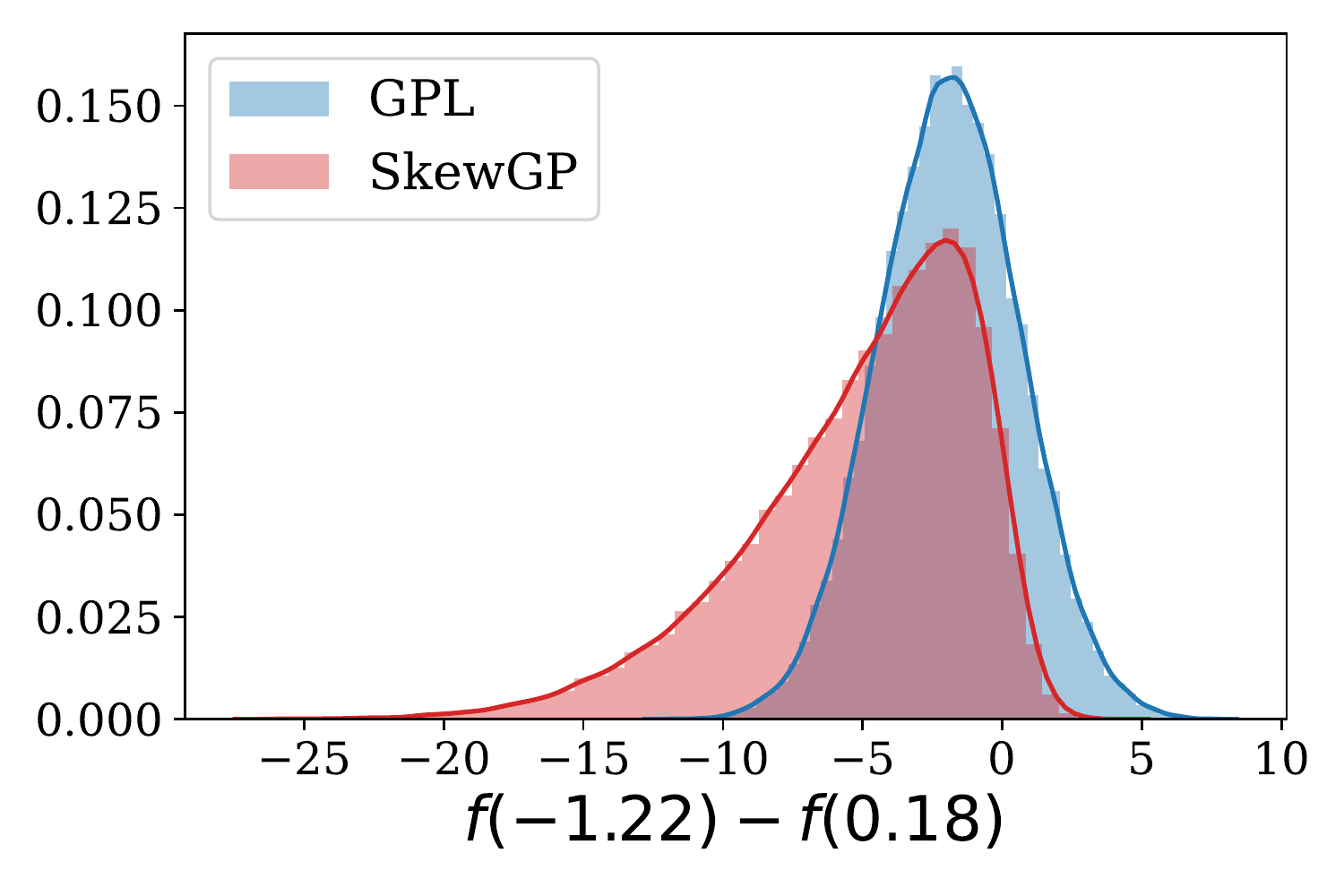} &
		\includegraphics[width=.42\linewidth]{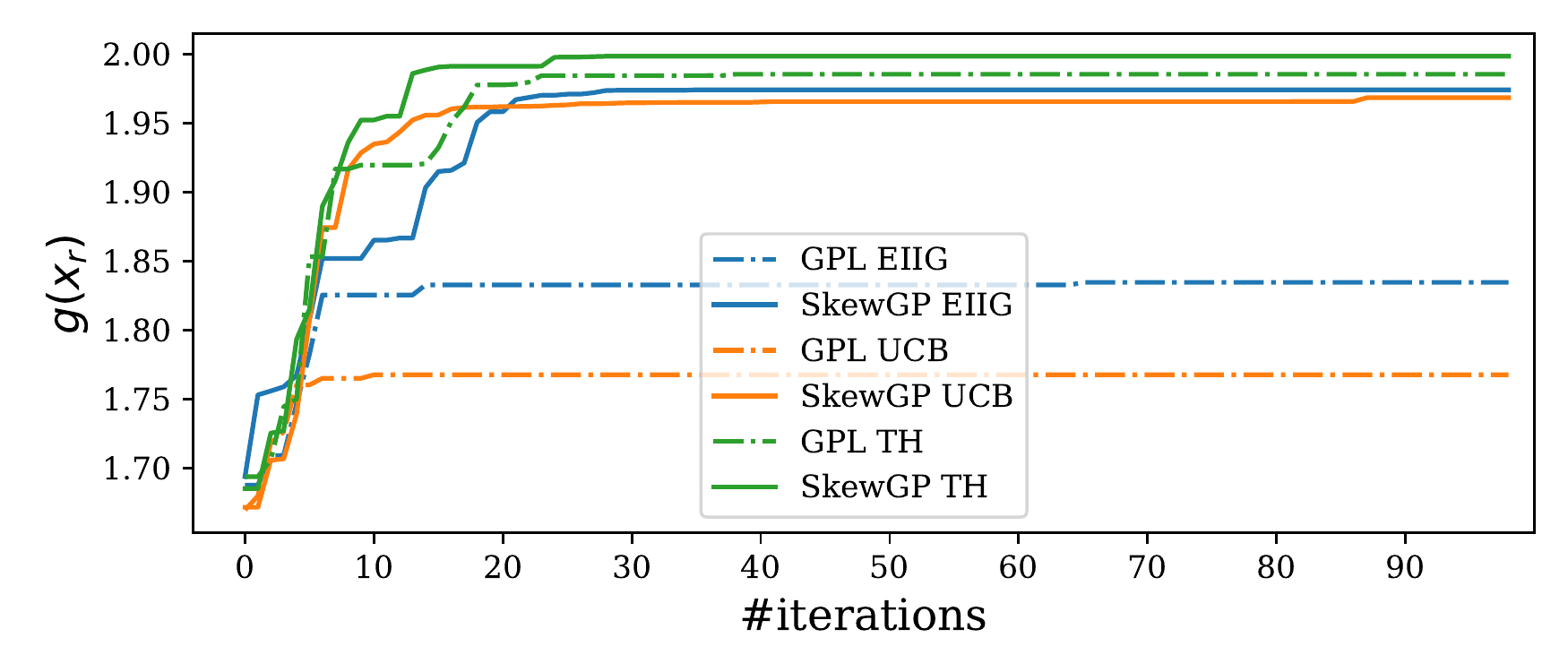}\\
		(a) & (b)\\
	\end{tabular}
	\caption{Left: skewed marginal. Right: convergence speed
		(EIIG $k=0.1$).}
	\label{Fig:figthird}
\end{figure}

Figure~\ref{fig:3} shows that the Laplace approximation for $f(x)-f(x_r)$ is much worse than SkewGP. 
Moreover, the posterior of $f(x)-f(x_r)$ is heavily skewed. The maximum magnitude of $SS(f(x)-f(x_r))$ is $-1.3$ (see the relative marginal posteriors in Figure \ref{Fig:figthird}(top)). 

Note from Figure \ref{fig:3}(1st-row, central) that, while the maximum of UCB for GPL and SkewGP almost coincides in the initial iteration, they significantly differ in the second iteration, Figure \ref{fig:3}(2nd-row, central): SkewGP's UCB selects a point very close to the global maximum, while GPL explores the border of the search space. This again happens in the subsequent two iterations, see \ref{fig:3}(3nd-row and 4th-row, central).

This behavior is neither special to this trial
nor to the UCB acquisition function, we repeat 
this experiment 20 times starting with $10$ initial (randomly selected) duels  and using all the three acquisition functions. 
We compare GPL versus SkewGP with fixed kernel hyperparameters (same as above) so that the only difference between the two algorithms is in the computation of the posterior.\footnote{In our implementation we compute the acquisition functions via Monte Carlo sampling (2000 samples).}
	We report the average (over the 20 trials) performance, which is defined as the value of $g$  evaluated at the current optimal point ${\bf x}_r$, considered to be the best by each method at each one of the 100  iterations.
	
The results are showed in Figure \ref{Fig:figthird}(bottom): SkewGP always outperforms
GPL. 
This is only due to  Laplace's approximation (hyper-parameters are the same).
In 1D, the differences are smaller for the Thompson (TH) acquisition function (due to the ``noise'' from  the random sampling step). However, SkewGP-TH converges faster.

%

\subsection{Optimization benchmarks}
We have considered for $g({\bf x})$ the same four benchmark functions used in \citep{gonzalez2017preferential}: `Forrester' (1D),  `Six-Hump
Camel' (2D), `Gold-Stein' (2D) and `Levy' (2D), and additionally
`Rosenbrock5' (5D) and `Hartman6' (6D). These are minimization problems.\footnote{We converted them into maximizations
	so that the acquisition functions in Section \ref{sec:acquisition} are well-defined.}
Each experiment starts with $10$ initial (randomly selected) duels and a total budget of $100$ duels are run.
Further, each experiment is repeated 20 times  with different initialization (the same for all methods) as in \citep{gonzalez2017preferential}. 
We compare PBO based on GPL versus SkewGP using the three different acquisition functions described in Section \ref{sec:acquisition}:
UCB, EIIG ($k=0.1$ and $0.5$) and Thompson (denoted as TH).
As before, 
we show plots of \#iterations versus $g({\bf x}_r)$.
In these experiments \textbf{we optimize the kernel hyperparameters 
by maximising the marginal likelihood for both GPL and SkewGP.}
\begin{figure}
	\centering
	\begin{tabular}{ll}
		\begin{minipage}{7cm}
			\includegraphics[height=3.0cm,trim={0.6cm 0.0cm 0.0cm 0.0cm }, clip]{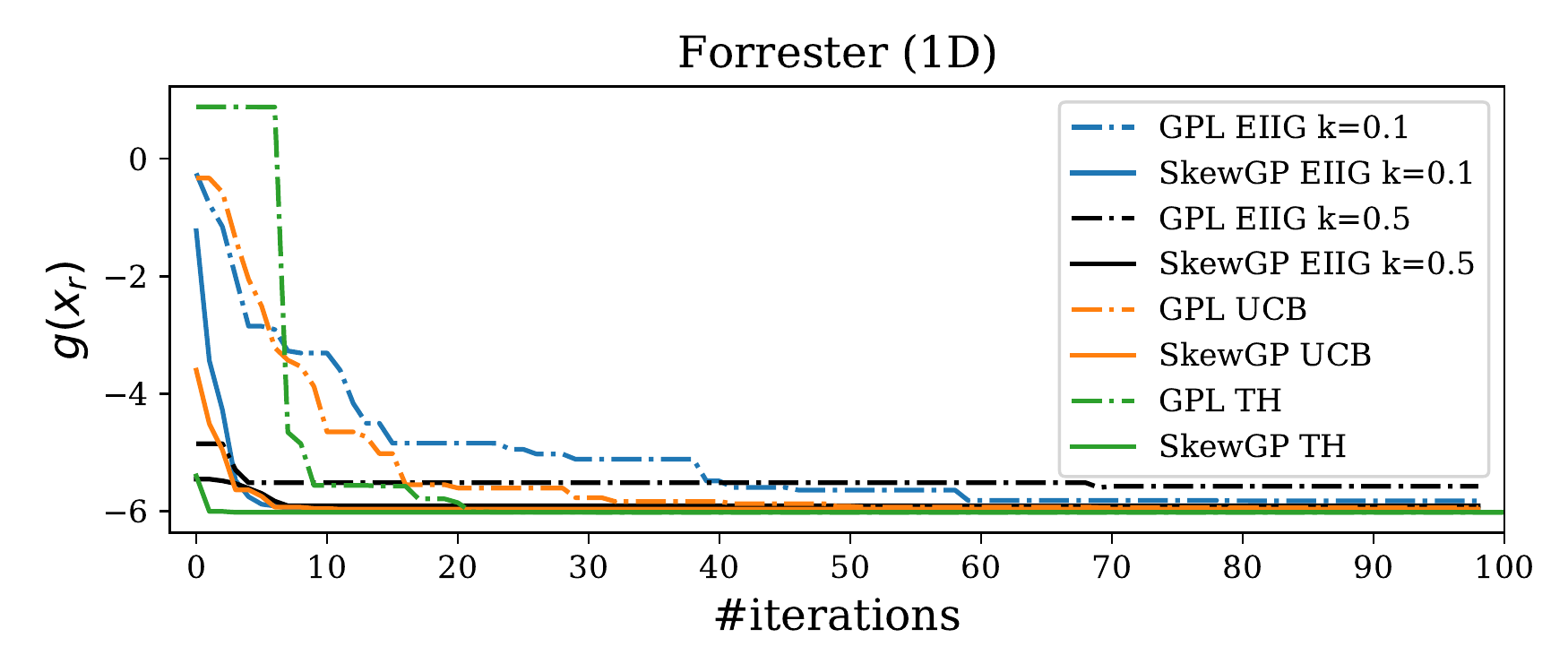}
		\end{minipage}& 
		\begin{minipage}{7cm}
			\includegraphics[height=3.0cm,trim={0.6cm 0.0cm 0.0cm 0.0cm }, clip]{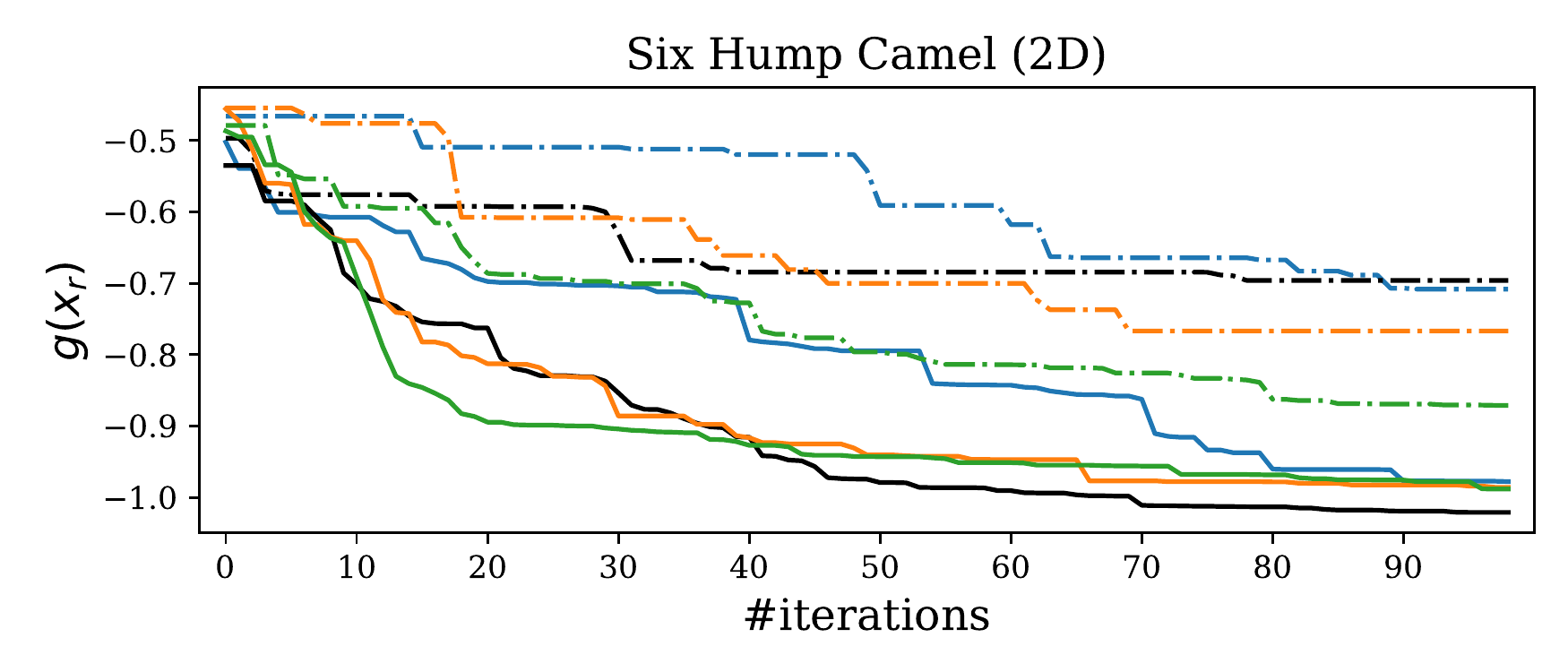}
		\end{minipage}\\
		\begin{minipage}{7cm}
			\includegraphics[height=3.0cm,trim={0.6cm 0.0cm 0.0cm 0.0cm }, clip]{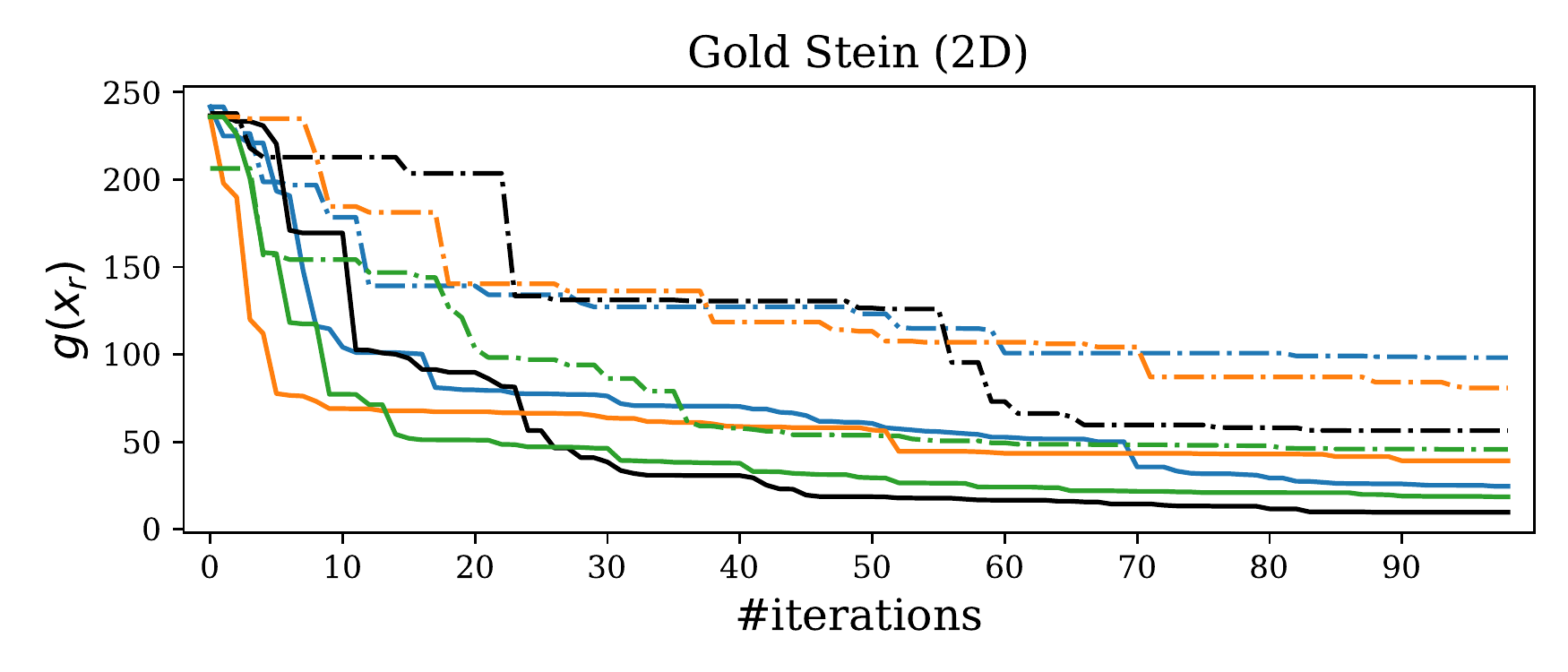}
		\end{minipage}& 
		\begin{minipage}{7cm}
			\includegraphics[height=3.0cm,trim={0.6cm 0.0cm 0.0cm 0.0cm }, clip]{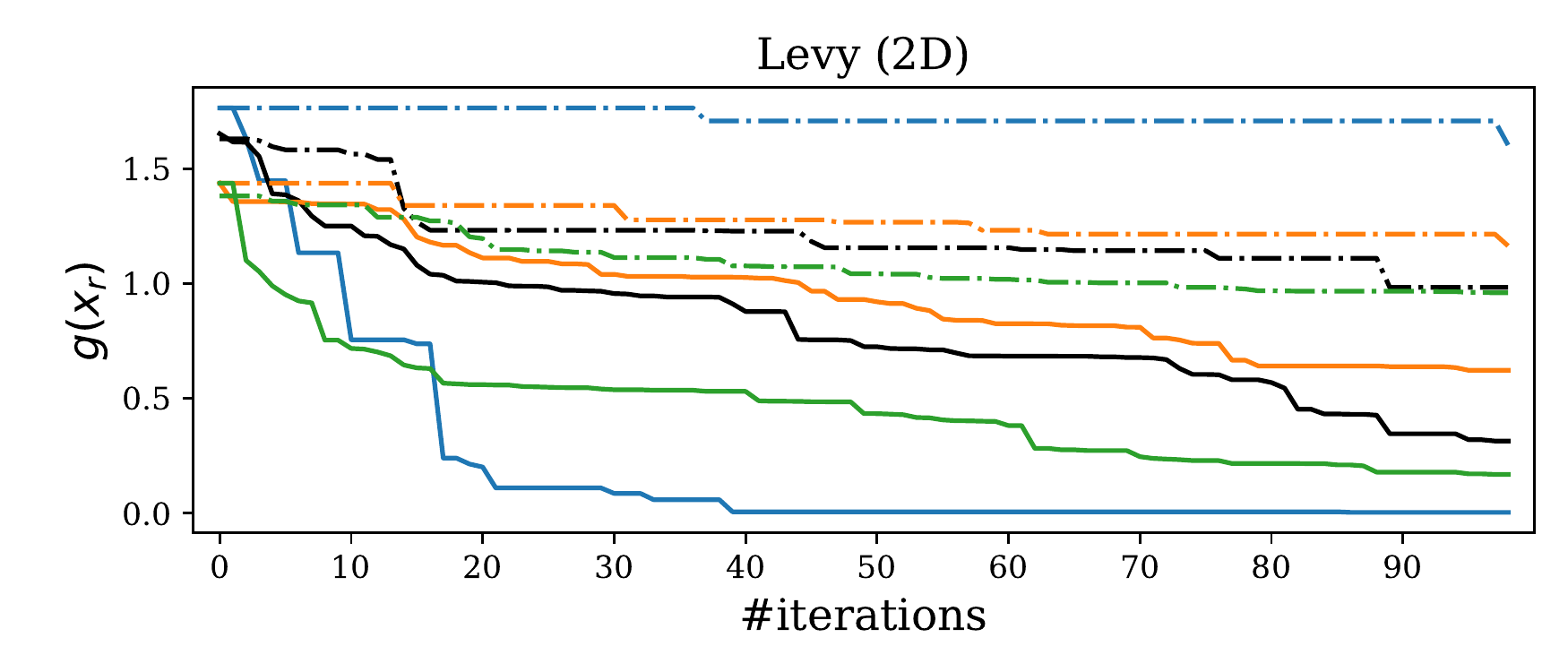}
		\end{minipage}\\
		\begin{minipage}{7cm}
			\includegraphics[height=3.0cm,trim={0.6cm 0.0cm 0.0cm 0.0cm }, clip]{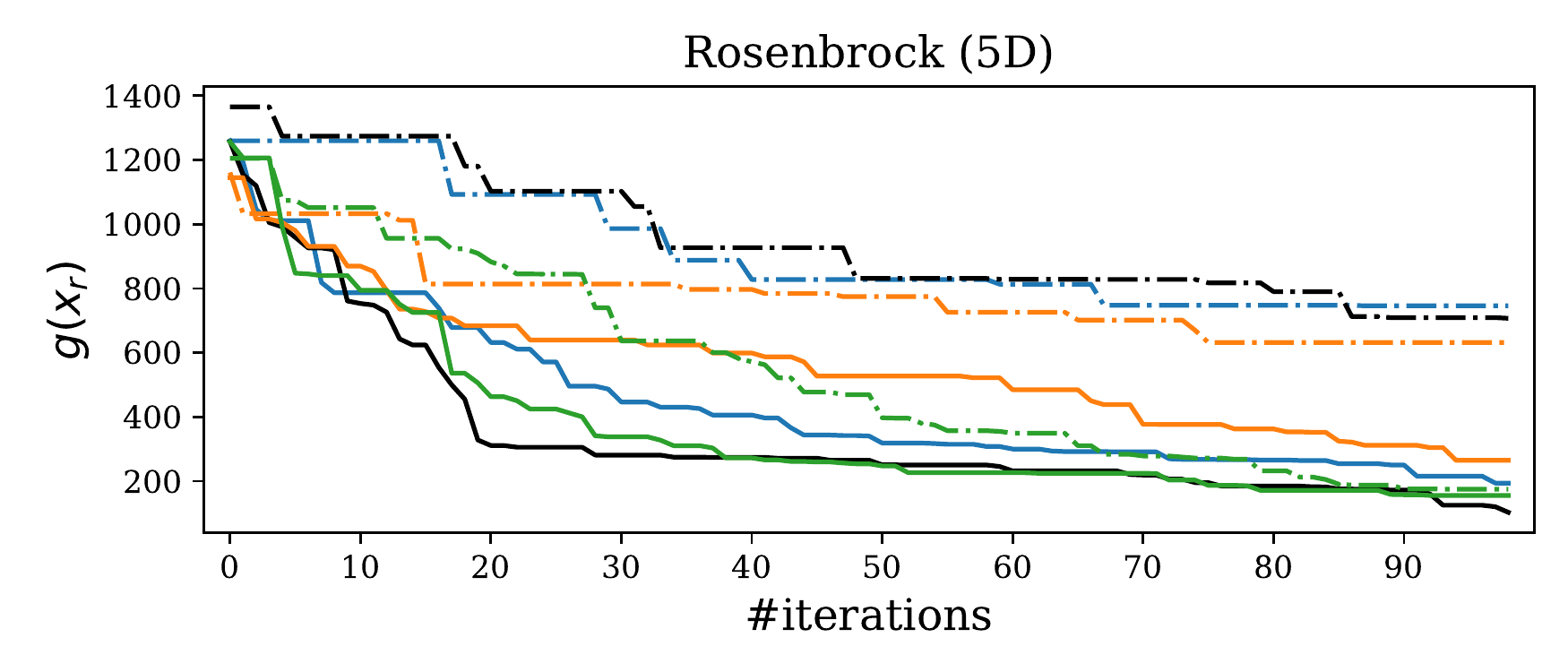}
		\end{minipage}& 
		\begin{minipage}{7cm}
			\includegraphics[height=3.0cm,trim={0.6cm 0.0cm 0.0cm 0.0cm }, clip]{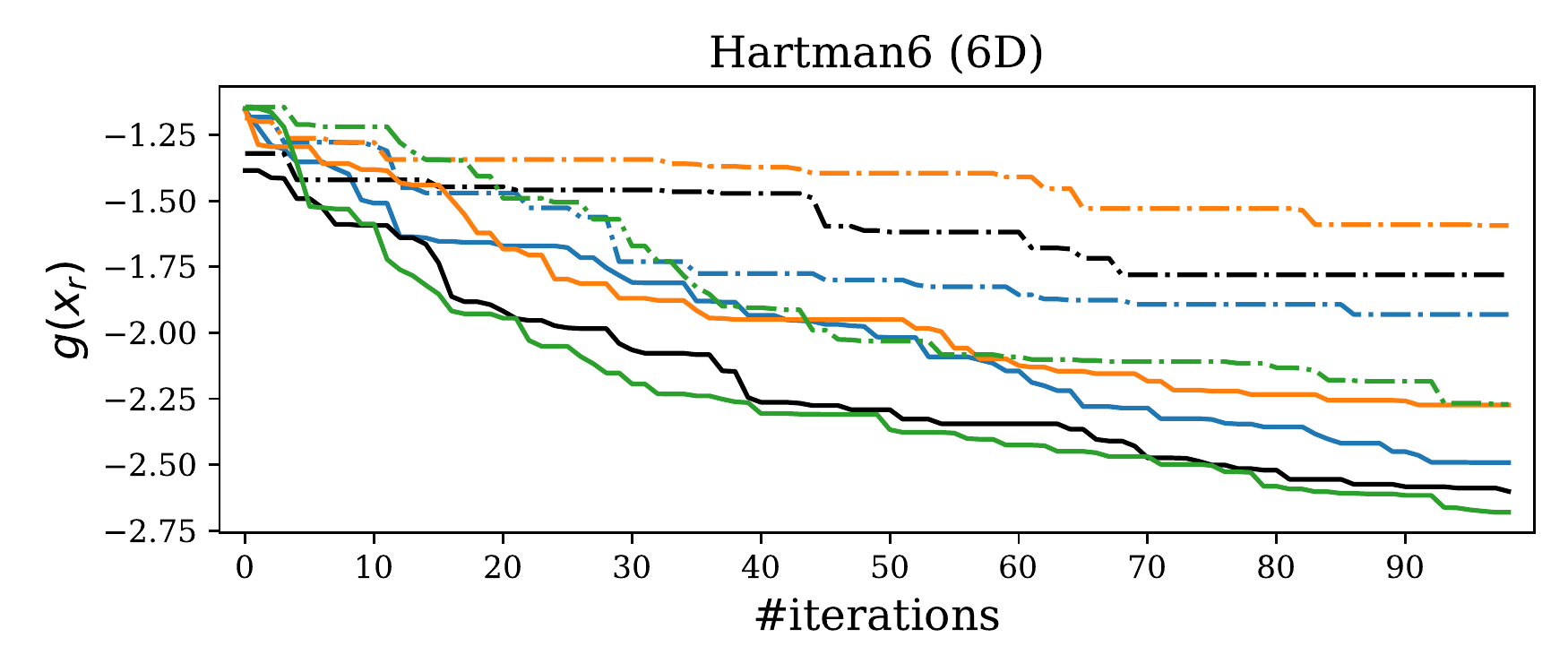}
		\end{minipage}\\
	\end{tabular}
	\rowcolors{2}{green!6}{white}  
	{\small
		\centering
		\begin{tabular}{c|cccccc}
			\rowcolor{green!20} 
			& Forrester & Six Hump Camel & Gold Stein & Levy & Rosenbrock5 & Hartman6 \\
			GPL PBO  & 557  &2986 & 3346 & 4704 & 4314 &5801 \\
			SkewGP PBO & 102 &2276 & 1430 & 1211 & 2615 &2178 \\
	\end{tabular}}
	\caption{Averaged results over 20 trials for GPL versus SkewGP on the 6 benchmark functions considering 3 different acquisition functions.  The x-axis represents the
		number of evaluation and the y-axis represents the value  of the true objective function at the current optimum ${\bf x}_r$. The table reports the median computational time per 100 iterations in seconds.}
	\label{fig:4}
\end{figure}
%
Figure \ref{fig:4} reports the performance of the 
different methods.
\textbf{Consistently across all benchmarks 
PBO-SkewGP outperforms PBO-GPL no matter the acquisition function.}
PBO-SkewGP has also a \textbf{lower computational burden} as showed
in the Table at the bottom of Figure \ref{fig:4} that compares
the median (over 80 trials, that is 20 trials times 4 acquisition functions) computational time per 100 iterations in seconds (on a standard laptop).

\subsection{Mixed preferential-categorical BO}

\begin{figure}[htp]
	\centering
	\begin{tabular}{ll}
		\begin{minipage}{15cm}
			\includegraphics[height=2.5cm,trim={0.0cm 0.0cm 0.0cm 0.0cm }, clip]{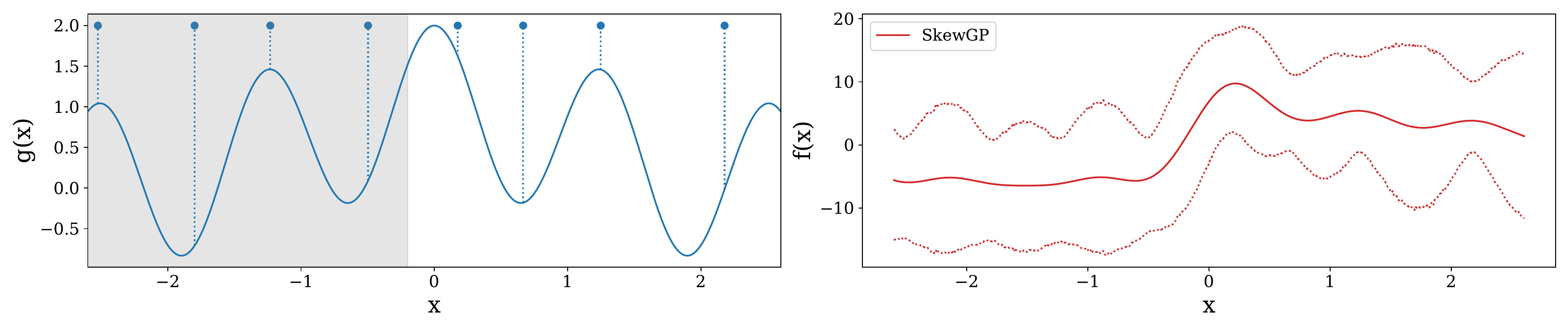}
		\end{minipage}
		
	\end{tabular}
	\caption{Mixed preference-classification BO}
	\label{fig:6}
\end{figure}

We examine now situations where certain unknown values of the inputs produce no-output preference and address it as a mixed preferential-categorical BO as described in Section \ref{subsec:mixed}.  
We consider again $g(x)=cos(5x)+e^{-\frac{x^2}{2}}$ and assume that any input $x\leq-0.2$ produces a non-valid output.
Figure  \ref{fig:6}(left) shows $g(x)$, the location of the queried points\footnote{The preferences are  $0.18\succ1.25$,
	$2.18\succ0.67$,
	$0.18\succ2.18$,
	$1.25\succ 0.67$,
	$0.18\succ0.67$.} and the no-valid zone (in grey).
Figure  \ref{fig:6}(right) shows the   predicted posterior preference function $f(x)$ (and relative 95\% credible region) computed according to SkewGP. We can see how SkewGP learns the non-valid region: the posterior mean is negative for $x\lesssim -0.2$  and positive otherwise (the oscillations of the mean for $x\gtrsim -0.2$ capture the preferences).
This is consistent with the likelihood in \eqref{eq:mixLikelihood}. 

\begin{figure}[htp]
	\centering
	\begin{minipage}{15cm}
		\centering
		\includegraphics[height=5cm,trim={0.0cm 0.0cm 0.0cm 0.0cm }, clip]{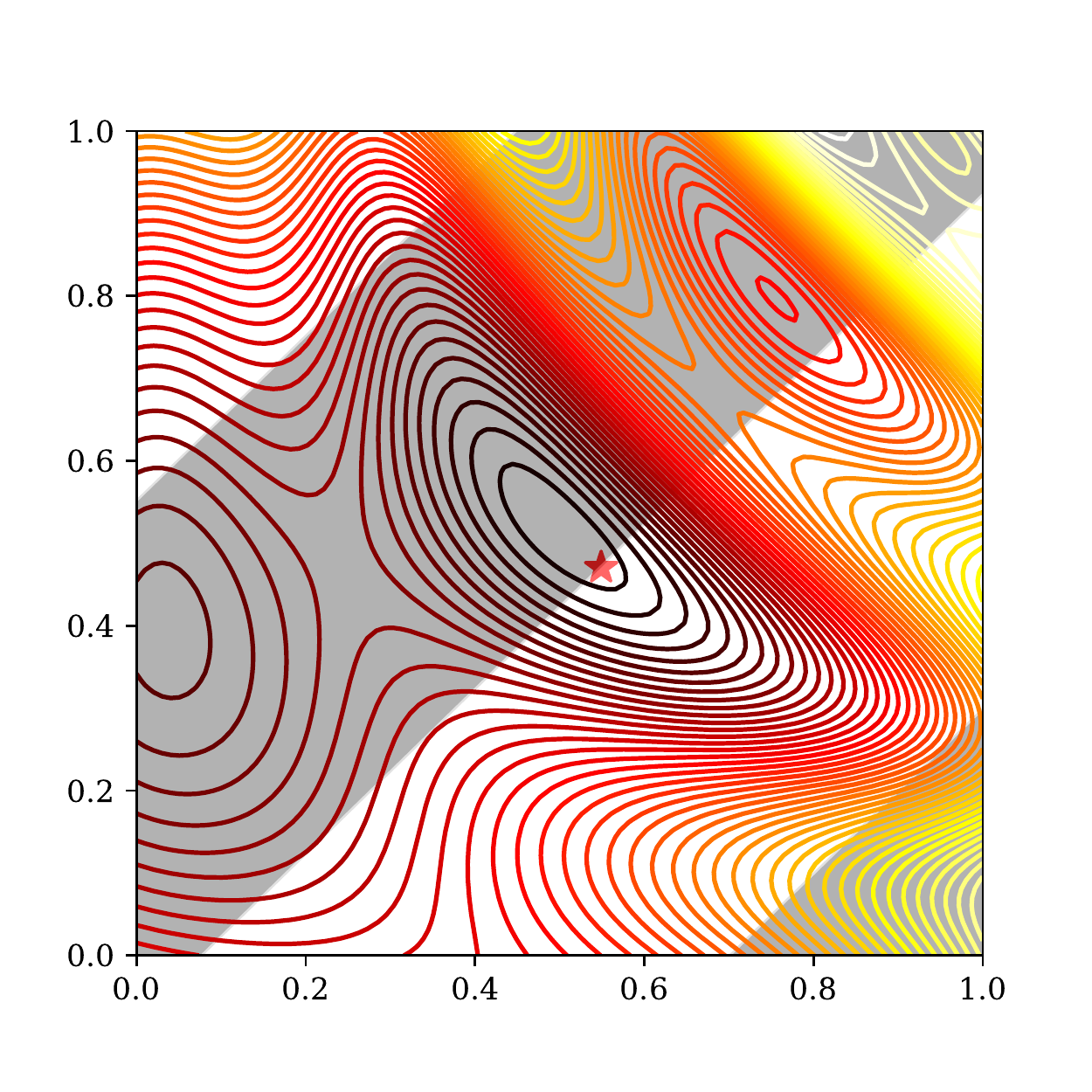}
		\includegraphics[height=3.6cm,trim={0.0cm 0.0cm 0.0cm 0.0cm }, clip]{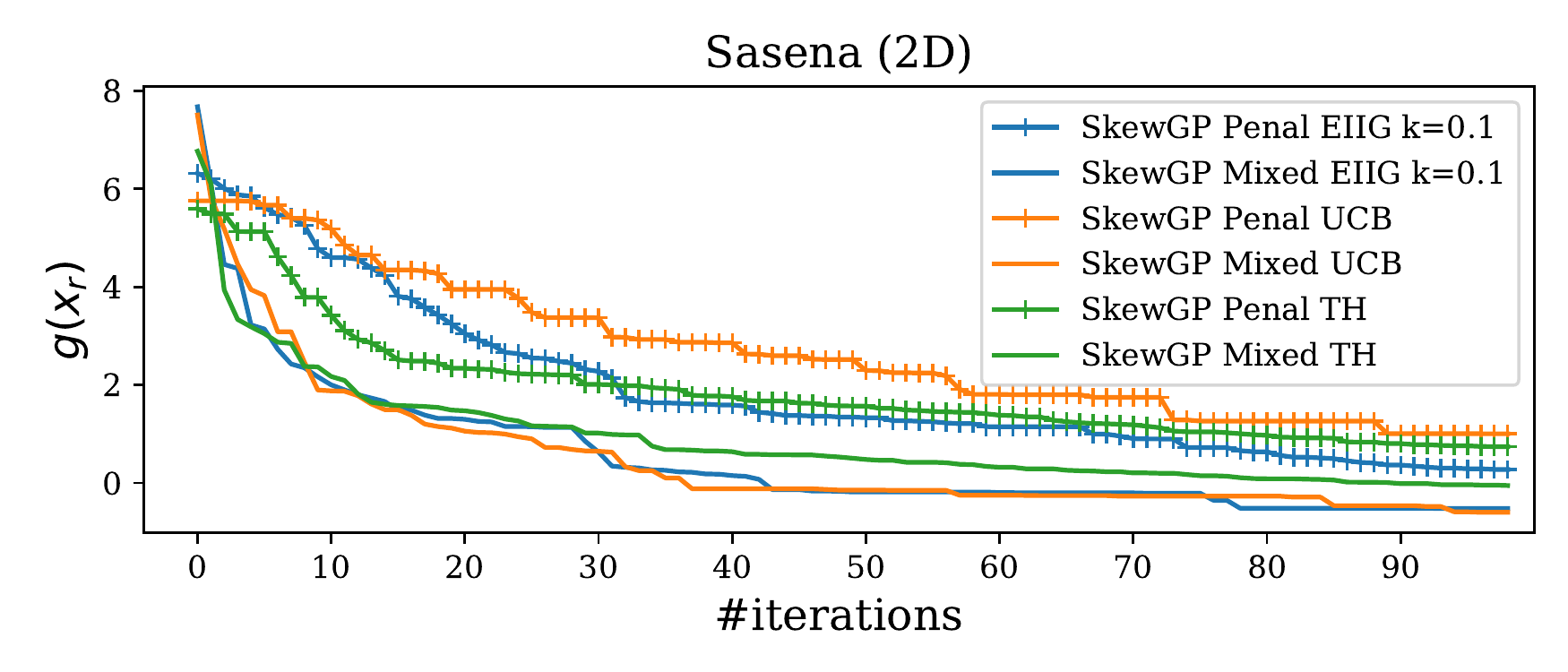}
	\end{minipage}
	\caption{Left: level sets of the benchmark function, 
		non-valid domain (grey bands) and location of the minimum (red star). Right: averaged results across 20 trials for
		SkewGP penalised PBO vs.\ SkewGP mixed preference-classification BO using 3 different acquisition functions.}
	\label{fig:7}
\end{figure}

We consider now the following benchmark optimization problem proposed in \citep{sasena2002exploration}:
$$
\min  \displaystyle{2+\tfrac{1}{100}(x_2-x_1^2)^2+(1-x_1)^2+2(2-x_2)^2+7\sin(\tfrac{1}{2}x_1)\sin(\tfrac{7}{10}x_1x_2)}
$$
with $0\leq x_1,x_2\leq 5$ 
and we assume that the input is \textit{valid} if $h({\bf x})=-\sin(x_1-x_2-\frac{\pi}{8})\leq 0$. We also assume that both the objective function and  $h({\bf x})$ are unknown. The goal is to find the minimum: $x^\star=[2.7450\ 2.3523]'$ with optimal cost $-1.1743$.
Figure \ref{fig:7}(left) shows the level curves
of the objective function, the non-valid zone (grey bands)
and the location  of the minimizer (red star).
We compare two approaches. The first approach uses  PBO-SkewGP  that  minimizes the objective  plus  a penalty term, $10^8 \max(0,h({\bf x}))^2$, that is non-zero in the \textit{non-valid} region. Adding a penalty for non-valid
inputs is the most common approach to deal with this type of problems. The second approach uses  a SkewGP based mixed preferential-categorical BO (SkewGP-mixed) that accounts for the \textit{valid}/\textit{non-valid} points as  in Section~\ref{subsec:mixed}.
Figure \ref{fig:7}(right) shows the performance of the two compared methods, which is consistent across the three
different acquisition functions: SkewGP-mixed converges  more quickly to the optimum.
This confirms that modelling directly this type of problems via the  mixed preferential-categorical likelihood in \eqref{eq:mixLikelihood}  enables the model to fully exploit the available information.
Also in this case, \textbf{SkewGP allows us to compute the corresponding posterior exactly.}

\section{Conclusions}

In this work we have shown that is possible 
to perform exact preferential Bayesian optimization by using Skew Gaussian processes. 
We have demonstrated that in the setting of preferential BO:
(i) the Laplace's approximation is very poor; (ii) given the strong skewness of the posterior,  any approximation of the  posterior that relies on a symmetric distribution will result to sub-optimal predictive performances and, therefore, slower convergence in PBO.
We have also shown that we can extend this model to deal with mixed preferential-categorical Bayesian optimisation, while still providing the exact posterior. We envisage an immediate extension of our current approach.
Many optimisation applications are  subject to safety constraints, so that inputs cannot be freely chosen from the entire input space. This leads to so-called safe Bayesian optimization \citep{gotovos2015safe}, that has been extended to safe preferential BO in  \citep{pmlr-v80-sui18a}. We plan to solve this problem exactly using SkewGP.

%
%
%

\appendix

\section{Background on the Skew-Normal distribution}
\label{app:Background}

In this section we provide more details on the skew-normal distribution. See, e.g.~\cite{azzalini2013skew} for a complete reference on skew-normal distributions and their parametrizations. 

\subsection{Additive representations}	
\label{sec:Additive}
The role of the latent dimension $s$ can be briefly explained as follows. 
Consider a random vector $\begin{bmatrix}
	\bx_0 \\
	\bx_1 
	\end{bmatrix} \sim N_{s+p}(0,M)$ with $M$ as in \eqref{eq:positivity} and define $\mathbf{y}$ as the vector with distribution $(\bx_1 \mid \bx_0+\bgamma>0)$. The density of $y$ can be written as 
	\begin{align*}
	f(\mathbf{y}) &= \frac{\int_{\bx_0+\bgamma>0}\varphi_{s+p}((\mathbf{t}_0,\mathbf{y});M)d\mathbf{\mathbf{t}_0}}{\int_{\bx_0+\bgamma>0}\varphi_s(\mathbf{t};\Gamma)d\mathbf{t}} = \varphi_p(\mathbf{y};\bar{\Omega})\frac{P(\bx_0+\bgamma>0 \mid \bx_1 = \mathbf{y})}{\Phi_s(\bgamma;\Gamma)} \\
	&= \varphi_p(\mathbf{y};\bar{\Omega})\frac{\Phi_s\left(\bgamma+\Delta^T\bar{\Omega}^{-1}\mathbf{y}; \Gamma-\Delta^T\bar{\Omega}^{-1}\Delta\right)}{\Phi_s(\bgamma;\Gamma)},
	\end{align*}
	where the first equality comes from a basic property of conditional distributions, see, e.g.\citep[Ch. 1.3]{azzalini2013skew}, and the second equality is a consequence of the multivariate normal conditioning properties. Then we have that $\bz = \bxi + \Domega\mathbf{y}\sim\text{SUN}_{p,s}(\bxi,\Omega,\Delta,\bgamma,\Gamma)$. 
	
	The previous representation provides an interesting point of view on the skew-Gaussian random vector, however the following representation turns out to be more practical for sampling from this distribution. Consider the independent random vectors 	
	$\mathbf{u}_0 \sim N_p(0, \bar{\Omega} - \Delta \Gamma^{-1}\Delta^T)$ and $\mathbf{u}_{1,-\bgamma}$, the truncation below $\bgamma$ of $\mathbf{u}_1\sim N_s(0,\Gamma)$. Then the random variable 
	\begin{equation*}
	\mathbf{z}_u = \bxi + \Domega(\mathbf{u}_0 + \Delta \Gamma^{-1}\mathbf{u}_{1,-\bgamma}),
	\end{equation*}
	is distributed as~\eqref{eq:sun}.
	\begin{proof}
		We can show that the representation $\bx_0,\bx_1$ is equivalent to $\mathbf{u}_0,\mathbf{u}_1$. Define $\mathbf{u}_1 = \bx_0$ and $\mathbf{u}_0 = \bx_1 - \mathbb{E}[\bx_1 \mid \bx_0]$, where $\begin{bmatrix}
		\bx_0 \\
		\bx_1
		\end{bmatrix} \sim N_{s+p}(0,M)$. Note that $\mathbb{E}[\bx_1 \mid \bx_0] = \Delta \Gamma^{-1}\bx_0$ and $\mathbf{u}_0 = \bx_1 - \Delta \Gamma^{-1}\bx_0 \sim N(0, \bar{\Omega} - \Delta \Gamma^{-1}\Delta^T)$. Then we have that $\mathbf{u}_0$ and $\mathbf{u}_1$ are independent. This can be verified by the fact that $\mathbf{u}_0$ and $\mathbf{u}_1$ are normally distributed with covariance $\operatorname{Cov}(\mathbf{u}_0,\mathbf{u}_1)=0$ which can be verified with algebraic computations. Finally note that $(\mathbf{u}_0 + \Delta \Gamma^{-1}\mathbf{u}_{1,-\bgamma})$ is distributed as $(\bx_1 \mid \bx_0 \bgamma>0)$.
	\end{proof}
	 The additive representation introduced above is used in Section~2.4 to draw samples from the distribution. 

\subsection{Closure properties}
\label{sec:closure}
The Skew-Normal family has several interesting properties, see \citet[Ch.7]{azzalini2013skew} for details. Most notably, it is close under marginalization and affine transformations. 
Specifically, if we partition $z = [z_1 , z_2]^T$,
where $z_1 \in \mathbb{R}^{p_1}$ and $z_2 \in \mathbb{R}^{p_2}$
with $p_1+p_2=p$, then
\begin{equation}
\label{eq:marginalFinDim}
\begin{array}{c}
z_1  \sim SUN_{p_1,s}(\bxi_1,\Omega_{11},\Delta_1,\bgamma,\Gamma), \vspace{0.2cm}\\
\text{with }~~
\bxi=\begin{bmatrix}
\bxi_1\\\bxi_2
\end{bmatrix},~~~
\Delta=\begin{bmatrix}
\Delta_1\\\Delta_2
\end{bmatrix},~~~
\Omega=\begin{bmatrix}
\Omega_{11} & \Omega_{12}\\
\Omega_{21} & \Omega_{22}
\end{bmatrix}.
\end{array}
\end{equation}
Moreover, \citep[Ch.7]{azzalini2013skew} the conditional distribution is a unified skew-Normal, i.e.,  
%
$$
(Z_2|Z_1=z_1) \sim SUN_{p_2,s}(\bxi_{2|1},\Omega_{2|1},\Delta_{2|1},\bgamma_{2|1},\Gamma_{2|1}),
$$
where 
\begin{align*}
\bxi_{2|1} & :=\bxi_{2}+\Omega_{21}\Omega_{11}^{-1}(z_1-\bxi_1), \quad
\Omega_{2|1} := \Omega_{22}-\Omega_{21}\Omega_{11}^{-1}\Omega_{12},\\
\Delta_{2|1} &:=\Delta_2 -\bar{\Omega}_{21}\bar{\Omega}_{11}^{-1}\Delta_1,\\
\bgamma_{2|1}& :=\bgamma+\Delta_1^T \Omega_{11}^{-1}(z_1-\bxi_1), \quad
\Gamma_{2|1}:=\Gamma-\Delta_1^T\bar{\Omega}_{11}^{-1}\Delta_1,
\end{align*}
and $\bar{\Omega}_{11}^{-1}:=(\bar{\Omega}_{11})^{-1}$.

In section~2.4 we exploit this property to obtain samples from the predictive posterior distribution at a new input $\bx^*$ given samples of the posterior at the training inputs.

\subsection{Sampling from the posterior predictive distribution}
\label{sec:samplingpost}
Consider a test point $\bx^*$ and assume we have a sample from the posterior distribution $f(X) \mid \mathcal{D}$. Consider the vector $\mathbf{\hat{f}} = [f(X)~~f(\bx^*)]^T$, which is distributed as $\text{SUN}_{n+1,s}(\hat{\bxi},\hat{\Omega}, \hat{\Delta}, \bgamma, \Gamma)$, where 
\begin{align*}
\hat{\bxi}=\begin{bmatrix}
\xi(X)\\\xi(\bx^*)
\end{bmatrix},~~~
\hat{\Delta}=\begin{bmatrix}
\Delta(X)\\\Delta(\bx^*)
\end{bmatrix},~~~
\hat{\Omega}=\begin{bmatrix}
\Omega(X,X) & \Omega(X,\bx^*)\\
\Omega(\bx^*,X) & \Omega(\bx^*,\bx^*)
\end{bmatrix}
\end{align*}
Then by using the marginalization property introduced above we obtain the formula in \eqref{eq:posteriorclass}.

\section{Proofs of the results in the paper}
\label{app:proofs}

\paragraph{Theorem~\ref{lemma:1}}
This proof is based on the proofs in \citep[Th.1 and Co.4]{durante2018conjugate}.
We aim to derive the posterior of $f(X)$.
The joint distribution of $f(X),\mathcal{D}$ is 
\begin{align}
p(\mathcal{D}|f(X))p(f(X))
=\Phi_m(Wf)\;\phi_n(f-\xi;\Omega)
\label{eq:numjoint}
\end{align}
where we have omitted the dependence on $X$ for easier notation. We note that 
$$
\begin{aligned}
\Phi_m(W f) 
=\Phi_m\left(W\xi+(\bar{\Omega}\Domega W^T)^{T}\bar{\Omega}^{-1}\Domega^{-1}(f-\xi); (W \Omega W^T + I_m)  - (W \Omega W^T)   \right)
\end{aligned}
$$ 
Therefore, we can write
\begin{align}
\nonumber
\Phi_m(Wf)\;\phi_n(f-\xi;\Omega) &=  \Phi_m\left(W\xi+(\bar{\Omega}\Domega W^T)^{T}\bar{\Omega}^{-1}\Domega^{-1}(f-\xi); (W \Omega W^T + I_m)  - (W \Omega W^T)   \right) \\
\nonumber
&\cdot \phi_n(f-\xi;\Omega) \\
\label{eq:prodcdf}
&= \Phi_m(m;M) \phi_n(f-\xi;\Omega)
\end{align}
with 
\begin{equation*}
m= W\xi+(\bar{\Omega}\Domega W^T)^{T}\bar{\Omega}^{-1}\Domega^{-1}(f-\xi)
\end{equation*}
and
\begin{equation*}
M= (W\Omega W^T+I_m)-W \Domega \bar{\Omega}\Domega W^T
\end{equation*}

From \eqref{eq:numjoint}--\eqref{eq:prodcdf} and the definition of the PDF of the SUN distribution, 
we can easily show  that we can rewrite \eqref{eq:numjoint} as a SUN distribution with updated parameters:
$$
\begin{aligned}
\tilde{\xi} & =\xi, \qquad \tilde{\Omega} = \Omega, \\
\tilde{\Delta} &=\bar{\Omega}\Domega W^T,\\
\tilde{\gamma}& =W\xi, \qquad
\tilde{\Gamma}= (W \Omega W^T + I_m).
\end{aligned}
$$

\paragraph{Theorem \ref{th:1}}

Consider the test point $\bx\in \mathbb{R}^d$ and the vector $\hat{f} = \begin{bmatrix}
f(X) \\
f(\bx)
\end{bmatrix} := [\mathbf{f}~~f_{*}]$ we have 
\begin{equation*}
p(\mathbf{f},f_{*}) = N\left(\begin{pmatrix}
\xi(X) \\
\xi(\bx)
\end{pmatrix}, \begin{pmatrix}
\Omega(X,X) & \Omega(X,\bx) \\
\Omega(\bx,X) & \Omega(\bx,\bx) \\
\end{pmatrix}\right)
\end{equation*}
and the predictive distribution is by definition
\begin{align*}
p(f_{*} \mid W) &= \int p(f_* \mid \mathbf{f}) p(\mathbf{f} \mid W)d\mathbf{f} \\
&=  \int p(f_* \mid \mathbf{f}) \frac{p(W \mid \mathbf{f})p(\mathbf{f})}{p(W)}d\mathbf{f} \\
&\propto \int p(f_*, \mathbf{f}) p(W \mid \mathbf{f})d\mathbf{f}
\end{align*}
We can then apply Lemma~\ref{lemma:1} with $\hat{f}$ and the likelihood $p([W \mid \mathbf{0}] \mid \hat{f}) = \Phi_{m+1}([W \mid \mathbf{0}] \begin{bmatrix}
f(X) \\
f(\bx)
\end{bmatrix}; I_m)$ which results in a posterior distribution 
\begin{equation*}
p\left(\begin{bmatrix}
f(X) \\
f(\bx)
\end{bmatrix} \mid [W\mid \mathbf{0}]\right) = \text{SUN}_{n+1,m}(\hat{\bxi},\hat{\Omega},\hat{\Delta},\hat{\gamma},\hat{\Gamma})
\end{equation*}
with 
\begin{align*}
\hat{\bxi} &= [\xi(X)~~\xi(\bx)]^T \\
\hat{\Omega} &= \begin{bmatrix}
\Omega(X,X) & \Omega(X,\bx) \\
\Omega(\bx,X) & \Omega(\bx,\bx) \\
\end{bmatrix} \\
\hat{\Delta} &= \begin{bmatrix}
\Omega(X,X) & \Omega(X,\bx) \\
\Omega(\bx,X) & \Omega(\bx,\bx) \\
\end{bmatrix}[W \mid \mathbf{0}]^T =\begin{bmatrix}
\Omega(X,X)W^T \\
\Omega(\bx,X)W^T
\end{bmatrix} \\
\hat{\gamma} &= [\xi(X)^T~~\xi(\bx)] \begin{bmatrix}
W^T \\
\mathbf{0}
\end{bmatrix} = \xi(X)^TW^T \\
\hat{\Gamma} &= [W \mid \mathbf{0}]\begin{bmatrix}
\Omega(X,X) & \Omega(X,\bx) \\
\Omega(\bx,X) & \Omega(\bx,\bx) \\
\end{bmatrix} \begin{bmatrix}
W^T \\
\mathbf{0}
\end{bmatrix} + I_{m} \\
&= W\Omega(X,X) W^T + I_{m}
\end{align*}

By exploiting the marginalization properties of the SUN distribution, see Section \ref{sec:closure}, we obtain
\begin{align}
\label{eq:marginalization}
p\left( f(\bx) \mid W,f(X) \right) = SUN_{1,m}\left(\xi(\bx),\Omega(\bx,\bx),\Omega(\bx,X)W^T, \xi(X)^TW^T, W\Omega(X,X)W^T +I_m\right).
\end{align}

\paragraph{Corollary~\ref{cor:Wpref}}

We can write the likelihood function as 
\begin{equation*}
p(\mathcal{D} \mid f(\mathcal{X})) = \Phi_m(U f(\mathcal{X}) - V f(\mathcal{X}); I_m ),
\end{equation*}
where $V \in \mathbb{R}^{m \times n}$ with $V_{i,j}=1$ if $v_i=x_j$ and $0$ otherwise and  $U \in \mathbb{R}^{m \times n}$ with $U_{i,j}=1$ if $u_i=x_j$ and $0$ otherwise. Then we can apply Lemma~\ref{lemma:1} for the posterior distribution of $f(X)$ and Theorem~\ref{th:1} for the posterior distribution of $f$ at an unobserved point.

\paragraph{Proposition~\ref{po:1}}
As described in Section \ref{sec:Additive}, if we  consider  a random vector $\begin{bmatrix}
 \bx_0 \\
 \bx_1 
 \end{bmatrix} \sim N_{s+p}(0,M)$ with $M=\left[\begin{smallmatrix}
\Gamma & \Delta \\
\Delta^T & \Omega
\end{smallmatrix}\right]
$ and define $\mathbf{y}$ as the vector with distribution $(\bx_1 \mid \bx_0+\bgamma>0)$, then it can be shown \citep[Ch. 7]{azzalini2013skew} that $\bz = \bxi + \Domega\mathbf{y}\sim\text{SUN}_{p,s}(\bxi,\Omega,\Delta,\bgamma,\Gamma)$.
This allows one to derive the following sampling scheme:
\begin{align}
\label{eq:sampling}
f &\sim {\bxi}+ \Domega\left(U_0+{\Delta}{\Gamma}^{-1}U_1\right),\\
\nonumber
U_0 &\sim \mathcal{N}(0;\bar{\Omega}-{\Delta}{\Gamma}^{-1}{\Delta}^T),\qquad 
U_1 \sim \mathcal{T}_{{\bgamma}}(0;{\Gamma}),
\end{align}
where $\mathcal{T}_{{\bgamma}}(0;{\Gamma})$ is the pdf of a multivariate Gaussian distribution truncated component-wise below $-{\bgamma}$. 
Then from the marginal \eqref{eq:marginalization} and the above sampling scheme (see Section \ref{sec:samplingpost}), we obtain the formulae in \eqref{eq:sampling}, main text.

\paragraph{Corollary \ref{co:ml}} This follows 
from the Theorem \ref{th:1}:
$\Phi_{m}(\tilde{\bgamma},\tilde{\Gamma})$ is the normalization constant of the posterior and, therefore, the marginal likelihood is
$
\Phi_{m}(\tilde{\bgamma},\tilde{\Gamma})$.
The lower bound was proven in \cite[Prop.2]{Benavoli_etal2020}
%

\section{Implementation}
\label{app:implementation}
\subsection{Laplace's approximation}
The  Laplace's approximation for preference learning was implemented as described in \cite{ChuGhahramani_preference2005}.
We use standard Bayesian optimisation to
optimise the hyper-parameters of the kernel by maximising
the Laplace's approximation of the marginal likelihood.

\subsection{Skew Gaussian Process}
To compute $\Phi_{|B_i|}(\cdot)$  in \eqref{eq:marginalLikelihood}, we use the routine
proposed in \cite{trinh2015bivariate}, that computes multivariate normal probabilities using bivariate conditioning.  This is very fast.
We optimise  the hyper-parameters of the kernel by maximising
the lower bound in \eqref{eq:marginalLikelihood}
and we use simulated annealing.

\subsection{Acquisition function optimisation}
In sequential BO, our objective is to seek a new data point ${\bf x}$ which will allow us to
get closer to the maximum of the target function $g$.
Since $g$ can only be queried via preferences, this is obtained by optimizing w.r.t.\ ${\bf x}$ a dueling acquisition function $\alpha({\bf x},{\bf x}_r)$, where ${\bf x}_r$ is the best point found so far, that is the point that has the highest probability of
winning most of the duels (given the observed data $\mathcal{D}$) and, therefore, it is the most likely point maximizing $g$.

For both the models (Laplace's approximation (GPL) and SkewGP) we compute the acquisition functions  $\alpha({\bf x},{\bf x}_r)$ via Monte Carlo sampling from the posterior.
In fact, although for GPL some analytical formulas are available (for instance for UCB), by using random sampling (2000 samples with fixed seed) for both GPL and SkewGP  \textbf{removes any  advantage of SkewGP over GPL due to the additional exploration effect of Monte Carlo sampling.}
Computing $\alpha({\bf x},{\bf x}_r)$ in this way is very fast for both the models (for SkewGP this is due to lin-ess).
We then optimize $\alpha({\bf x},{\bf x}_r)$: (i) by computing  $\alpha({\bf x},{\bf x}_r)$  for 5000 random 
generated value of ${\bf x}$ (every time we use  the same random points for both SkewGP and GPL); (ii) the best random instance is then used as initial guess for L-BFGS-B.

%
%

\bibliographystyle{apalike}
\bibliography{biblio}

\end{document}